\documentclass{article}

\usepackage{arxiv}

\usepackage[utf8]{inputenc} 
\usepackage[T1]{fontenc}    
\usepackage{hyperref}       
\usepackage{url}            
\usepackage{longtable}
\usepackage{booktabs} 
\usepackage{ragged2e} 
\usepackage{tabularx}
\usepackage{amsfonts}       
\usepackage{nicefrac}       
\usepackage{microtype}      
\usepackage{lipsum}
\usepackage{graphicx}
\graphicspath{ {./images/} }

\title{Agentic AI Governance and Lifecycle Management in Healthcare}

\author{
 Chandra Prakash \\
  School of Computer Information Sciences\\
  University of the Cumberlands\\
  Williamsburg 40769, Kentucky, United States \\
  \texttt{cprakash@outlook.com} \\
   \And
 Mary Lind \\
  School of Computer Information Sciences\\
  University of the Cumberlands\\
  Williamsburg 40769, Kentucky, United States \\
  \texttt{mary.lind@gmail.com} \\
  \And
 Avneesh Sisodia \\
  School of Computer Information Sciences\\
  University of the Cumberlands\\
  Williamsburg 40769, Kentucky, United States \\
  \texttt{a.sisodia.k@gmail.com} \\
}

\begin{document}
\maketitle
\begin{abstract}
\textbf{Background:} Healthcare organizations are beginning to embed agentic AI into routine workflows, including clinical documentation support and early-warning monitoring. As these capabilities diffuse across departments and vendors, health systems face agent sprawl, causing duplicated agents, unclear accountability, inconsistent controls, and tool permissions that persist beyond the original use case. Existing AI governance frameworks emphasize lifecycle risk management but provide limited guidance for the day-to-day operations of agent fleets. \textbf{Objective:} We developed a Unified Agent Lifecycle Management (UALM) framework, a five-layer governance architecture with a companion maturity model featuring measurable KPI-linked thresholds, and evaluated its expected operational behavior through Monte Carlo simulation. \textbf{Methods:} We conducted Monte Carlo simulations using three synthetic healthcare organization profiles (small, medium, large) under four governance conditions: no governance (baseline), registry-only, NIST AI RMF-Lite, and UALM. All event probabilities, detection rates, and adoption curves were based on assumptions and tested through sensitivity analyses. Seven KPIs were tracked over six monthly time steps with stochastic event injection. Statistical comparisons used Mann-Whitney U tests, Cohen’s d effect sizes, and 95\% confidence intervals. Given multiple pairwise comparisons, adjusted significance thresholds were applied using the Holm-Bonferroni correction. Sensitivity analyses examined parameter robustness and the dependency of the governance ramp-up curve. \textbf{Results:} UALM showed consistently favorable effect sizes across KPIs; statistical significance was observed across pairwise comparisons. UALM reduced incident rates by 56-63\% versus no governance across all scenarios. Against the most relevant comparator (NIST RMF-Lite), UALM demonstrated large effect sizes (Cohen’s d = 0.99-6.06) on all KPIs, with the greatest advantages in credential revocation (-80\%, d = 6.06), tool-call logging (+57\%, d = 2.95), and PHI minimization (+77\%, d = 2.83). Only UALM-governed organizations achieved Level 2 maturity (18-21\% of runs), providing preliminary evidence that the maturity model can discriminate among governance conditions. Sensitivity analysis showed UALM’s advantage was robust across parameter variations (56-63\% incident reduction even under worst-case conditions), but it was dependent on sustained adoption commitment; stalled governance ramp-up erased approximately 80\% of the benefits. \textbf{Conclusions:} UALM maps recurring gaps onto five control-plane layers: (1) an identity and persona registry, (2) orchestration and cross-domain mediation, (3) PHI-bounded context and memory, (4) runtime policy enforcement with kill-switch triggers, and (5) lifecycle management and decommissioning linked to credential revocation and audit logging. UALM is projected to provide agent-specific governance advantages beyond what generic AI risk management offers under the modeled assumptions. The framework’s value is greatest in its lifecycle automation, policy enforcement, and PHI protection capabilities, which are not explicitly represented in the registry-only and NIST RMF-Lite comparator models. However, centralization overhead scales non-linearly with fleet size, and sustained organizational commitment is essential for realizing benefits. The companion maturity model and UALM support staged adoption, offering healthcare CIOs, CISOs, and clinical leaders an implementable pattern for audit-ready oversight that preserves local innovation and enables safer scaling across clinical and administrative domains.

\end{abstract}


\section{Introduction}
During 2024–2025, healthcare organizations began moving from generative AI pilots toward more autonomous, goal-directed systems commonly described as agentic AI \cite{Brodeur2025} \cite{Karunanayake2025}. While 2023 emphasized text generation and chatbots, the focus has shifted to proactive, goal-directed systems known as Agentic AI. Agentic AI can function as a workflow participant capable of planning, tool use, and goal-directed execution \cite{Shokrollahi2025} \cite{Zhang2023}. Unlike traditional models that wait for human queries, Agentic AI autonomously monitors data, reasons through clinical contexts, and executes multi-step actions \cite{Banerjie2025}. The healthcare industry has started to see scaled implementation, with peer-reviewed studies and clinical pilot programs increasing between 2024 and 2025 \cite{Gorenshtein2025}. The emergence of Multimodal Large Language Models (MLLMs) like Med-PaLM 2 and Med-Flamingo has facilitated this shift, allowing for the synthesis of diverse data sources, including medical imaging and real-time information \cite{Karunanayake2025}. The deployment of real-time documentation tools like Nuance DAX Copilot has been associated with reductions in documentation burden \cite{Liu2024}, while ICU-based agents monitor vital signs to alert teams in experimental or pilot contexts, to sepsis or arrhythmias before they become symptomatic \cite{Liu2025}. The future growth of Agentic AI is defined by a shift from reactive to proactive care, where autonomous systems intervene before a patient’s condition deteriorates \cite{Karunanayake2025} \cite{Liu2025}. However, without proper design-time limits and runtime governance, the proliferation of autonomous agents, known as agent sprawl, can lead to inefficiencies and safety risks \cite{Banerjie2025}. Research indicates an inverted-U relationship between the number of agents and task performance\cite{Gorenshtein2025}. A small set of specialized agents can improve performance; however, teams larger than ~4–5 agents may degrade outcomes due to coordination overhead \cite{Gorenshtein2025}. Ambiguous ownership, drifting controls, duplicated capabilities, and an expanding attack surface can lead to safety and compliance risks over time. Consequently, the challenge is not only improving care, but governing agent sprawl so that autonomy does not create redundancy, accountability gaps, or unmanaged risk. 

\subsection{Background}
The next generation of health systems is beginning to see AI evolve from “prediction support” to operational capacity. Healthcare Agentic AI systems have moved beyond “chatbot experiments” to plan steps, utilize tools such as APIs, databases, and workflow engines, and maintain context across various stages \cite{Schick2023}. Tool integration and context management enable agents to operate more like semi-autonomous workflow components \cite{OWASP}. For health system leaders, this shift is significant due to the cross-domain nature of operations. An eligibility check can initiate a claims pathway that intertwines with processes such as utilization management and patient experience. Introducing agentic capabilities separately across domains can lead to “Agent Sprawl,” resulting in overlapping agents, duplicated efforts, inconsistent controls, and unclear ownership. Local optimizations may cause system-wide friction and prolonged identities that outlast their original purpose. The autonomy of these systems alters the leadership risk profile by introducing behavioral uncertainty and broadening the impact of one agent’s actions on another’s processes. Past issues with prompt injection illustrate the potential for security vulnerabilities in LLM applications \cite{Prakash2026, Nolan2025}. These risks are especially relevant when agents have tool access, memory, and permissioned workflow actions. Governance also becomes complex, as misalignments may not be evident during routine evaluations but may surface upon deployment \cite{OpenAI2025}. The autonomy and governance issues challenge traditional assurance processes, underscoring the need to monitor for covert, goal-directed behaviors rather than focusing solely on accuracy \cite{OpenAI2025, Greenblatt2024}. \\

In healthcare, these risks escalate due to the handling of PHI and critical workflows \cite{HHS2009}. Although HIPAA does not prescribe agent-specific controls, its access control, audit control, integrity, and transmission security requirements create governance expectations for systems handling PHI. 

\subsection{Prior Work}
Table \ref{tab:govlife} summarizes the current state of AI agent governance and lifecycle management standards. The reviewed literature and guidance suggest that, though principles are available, lifecycle control planes are underspecified. Agent registration and discovery mechanisms are becoming mainstream, but governance enforcement remains disjointed. There is also a recognition of growing attack surface through prompt injection, insecure output handling, tool misuse, and privilege abuse. There are occasional discussions about agent sprawl, but there is minimal discussion on governance and lifecycle management in AI agent fleets.

\begin{longtable}{p{2.0cm}p{2.0cm}p{3.0cm}p{2.0cm}p{1.5cm}p{3.2cm}p{1.0cm}}
    \caption{Reviewed Frameworks and their Limitations}
    \label{tab:govlife} \\
        \toprule
        \textbf{Prior Work} & \textbf{Category} & \textbf{Contribution} & \textbf{Lifecycle Coverage} & \textbf{Conflict handling} & \textbf{Key Limitations} & \textbf{Refs} \\ 
        \midrule
        \endfirsthead
        \multicolumn{7}{c}{\small \textit{Table \ref{tab:govlife} (continued)}}\\
        \toprule
        \textbf{Prior Work} & \textbf{Category} & \textbf{Contribution} & \textbf{Lifecycle Coverage} & \textbf{Conflict handling} & \textbf{Key Limitations} & \textbf{Refs} \\ 
        \midrule
        \endhead
        
        \midrule
        \multicolumn{7}{r}{\small \textit{Table continues on next page}}\\
        \endfoot
        
        \bottomrule
        \endlastfoot
        
        EU AI Act & Regulatory & Risk-tiering, governance expectations & Org-level governance; compliance controls around development + deployment & Legal recourse, not technical runtime dispute resolution & Not tailored to fast-evolving multi-agent behaviors; compliance can lag operational realities & \cite{Commission2025} \\
        \midrule
        NIST AI RMF 1.0 & Risk mgmt framework & GOVERN-MAP-MEASURE-MANAGE functions to structure policies, controls, metrics, and monitoring for trustworthy AI systems. & End-to-end risk management & Indirect (process-based), not an inter-agent conflict protocol & agent-specific operationalization, such as identity, discovery, runtime authorization, and audit. & \cite{Tabassi2023} \\
        \midrule
        GenAI Profile for AI RMF & Risk mgmt. profile & Adds GenAI-specific risk considerations and actions aligned to AI RMF & Strong on lifecycle risk actions + continuous monitoring & Indirect & Not a dedicated agent lifecycle standard; organizations still must design registry, policy enforcement, deprovisioning, and runtime guardrails. & \cite{NIST2024} \\
        \midrule
        AI TRiSM & Governance approach & Consolidates trust/risk/security practices around explainability, ModelOps, security, privacy, and governance as an enterprise management lens. & Strong for monitoring + operational governance & Indirect & Lacks consensus benchmarks and agent-specific mechanisms. & \cite{Habbal2024} \\
        \midrule
        SAGA & Security architecture & Architecture for user-controlled agent lifecycle & Registration, identity, authorization, policy updates, revocation & Implicit & Enterprise-scale federation and multi-provider trust are still hard & \cite{Syros2025} \\
        \midrule
        ETHOS & Decentralized governance & Proposes blockchain / smart-contract-based registry + compliance artifacts and decentralized dispute ideas for agent ecosystems. & Registry + compliance attestation; conceptual lifecycle governance & Explicit & Heavy coordination; risk-tiering may oversimplify; adoption barriers & \cite{Chaffer2025} \\
        \midrule
        NANDA Index & Discovery / identity layer & Internet-scale discovery / identifiability/ authentication for agents beyond DNS assumptions; emphasizes verifiable “AgentFacts” style discovery. & Discovery, identity, and authentication primitives & NA & Strong on “find/verify agents,” weaker on enterprise policy enforcement, lifecycle controls, and cross-domain governance workflows. & \cite{Raskar2025} \\
        \midrule
        AGNTCY Agent Directory Service (ADS) & Registry / interoperability & Distributed directory for agent capability metadata + provenance & Discovery + provenance; supports ecosystem-scale inventory & NA & Policy enforcement is out of scope. & \cite{Muscariello2025} \\
        \midrule
        Google Agent2Agent & Interoperability / discovery & Standardizes how agents advertise identity/capabilities (“Agent Cards”) and communicate across enterprise estates; useful substrate for governance inventories. & Discovery + interoperability (interface-level) & NA & Governance standards, such as policy enforcement, audit, and lifecycle control, must be layered on & \cite{Surapaneni2025} \\
        \midrule
        SD-JWT / SD-JWT VC & Privacy-preserving credentials & Selective disclosure of claims supports privacy-preserving agent identity/capability attestations across domains. & Identity / attestation in discovery + access decisions & NA & Credential standards do not define enterprise policy semantics, revocation processes, or runtime decision control planes by themselves. & \cite{Nandakumar2025} \\
        \midrule
        OWASP Top 10 for LLM & Risk taxonomy & Codifies common failure modes (prompt injection, insecure output handling, supply chain, etc.) and extends to agentic-specific risks (goal hijack, tool misuse, identity/privilege abuse, etc.) for governance requirements. & Strong for threat modeling + control requirements across build/run & Indirect & Defines taxonomies but does not offer a unified lifecycle governance architecture. & \cite{OWASP} \\
        \midrule
        Agentic Profiles & Risk characterization for governance & Defines four dimensions, autonomy, efficacy, goal complexity, and generality, to tailor governance intensity and controls to agent capability profiles. & Helps decide governance “tiers” and supervisory needs across the lifecycle & Indirect & Needs mapping to implementable enterprise control sets. & \cite{Kasirzadeh2025}  \\
        \midrule
        Anthropic’s Alignment faking & Deceptive behavior risk & Empirical evidence that models can appear aligned while preserving hidden objectives motivating continuous monitoring, robust oversight, and auditability in agent operations. & Emphasizes runtime monitoring + evaluation & Indirect & Does not provide lifecycle architecture.  & \cite{Greenblatt2024}  \\
        \midrule
        Scheming evaluations & Deceptive behavior risk & Methods to detect/reduce hidden-misalignment behaviors in controlled tests; informs “safety gates” for agents. & Build-time + runtime evaluation/monitoring & Indirect & Not an enterprise lifecycle standard. & \cite{OpenAI2025}  \\
        
\end{longtable}

\subsection{The Need for Agentic AI Governance and Lifecycle Management}
AI governance frameworks such as NIST’s AI Risk Management Framework \cite{NIST2024} and the EU AI Act \cite{Commission2025} increasingly emphasize continuous, lifecycle-based risk management, yet they lack specifics on operationalizing these for Agentic AI systems. In healthcare organizations, there remains limited operational guidance of a minimally implementable agent lifecycle control plane governing how agents are registered, authorized, monitored, updated, and retired. This makes it challenging to sustain compliance and operational safety at scale \cite{Commission2025, NIST2025}. \\
Ongoing interoperability, discovery efforts, and privacy-preserving credentials promise improved visibility and trust signaling. However, they are rarely coupled with end-to-end policy enforcement across organizational boundaries \cite{Surapaneni2025, Nandakumar2025}. Behavioral evidence also suggests misalignment can be strategic and evade static testing, reinforcing the need for telemetry-driven governance that adjusts oversight for an agent’s autonomy, generality, goal complexity, and clinical or operational criticality\cite{OpenAI2025, Greenblatt2024}. These gaps become severe under HIPAA regulations that require demonstrable access controls and audit controls for systems handling PHI. However, agent-specific, audit-ready operational artifacts, such as registry fields, provenance, policy decision records, tool-call logs, and deprovisioning evidence, are not standardized. A lack of conflict resolution and of measurable indicators of agent sprawl remain under-modeled, leaving healthcare organizations without clear indicators of security, compliance, and operational risk \cite{HHS2009} as agent fleets grow.

\subsection{Research Questions}
\textbf{RQ1:} What lifecycle control-plane capabilities are needed to govern agentic AI sprawl in healthcare?
\textbf{RQ2:} Under modeled assumptions, how does UALM compare with no-governance, registry-only, and NIST RMF-Lite governance conditions across operational KPIs?

\subsection{Contribution}
This study makes three contributions. First, we present the Unified Agent Lifecycle Management (UALM) framework, a five-layer governance architecture derived from practice-oriented synthesis of governance standards, agent security literature, and healthcare compliance requirements. Second, we refine the companion maturity model with measurable, KPI-linked thresholds that enable reliable self-assessment. Third, we evaluate UALM through simulation-based analysis across four governance conditions, estimating its comparative performance under transparent parameter assumptions.

\section{Framework Development Approach}
To develop the proposed governance and lifecycle layers, we used a rapid, practice-oriented synthesis rather than a full systematic review. We first assembled the evidence base for Table \ref{tab:govlife} using a purposive search of (a) peer-reviewed literature on multi-agent systems and LLM/agent safety, (b) healthcare and enterprise security/governance guidance, and (c) widely cited industry frameworks on AI risk and operational control. We prioritized sources that described governance mechanisms, such as identity, access control, monitoring, auditability, change control, lifecycle processes, onboarding, versioning, retirement, or failure modes that arise when autonomous tools are connected to real workflows. Each source was reviewed using a simple extraction template that assessed the problem addressed, the proposed control mechanism, lifecycle coverage, and healthcare relevance. 
We then conducted an iterative thematic mapping exercise: recurrent gaps observed across sources, for example, weak ownership, inconsistent enforcement, and unclear decommissioning, were translated into control-plane requirements and grouped into a small number of layers that together form an end-to-end operating model. The layering was refined through repeated internal review focused on leadership questions, who owns decisions, what gets enforced, what is auditable, and what triggers shutdown or retirement. We also double-checked the draft model through informal conversations with healthcare IT/security leaders and incorporated feedback; these discussions were used only to stress-test practicality, no claims of qualitative saturation or thematic generalizability are made.

\section{Unified Agent Lifecycle Management: A Five-Layer Control-Plane Framework}
Before proposing the blueprint for the governance and lifecycle framework, we first define a maturity model to make governance progression measurable, identify gaps, and provide a roadmap for improvement. 
\subsection{Maturity Model}
\begin{figure}[!h]
    \centering
    \includegraphics[width=1.0\linewidth]{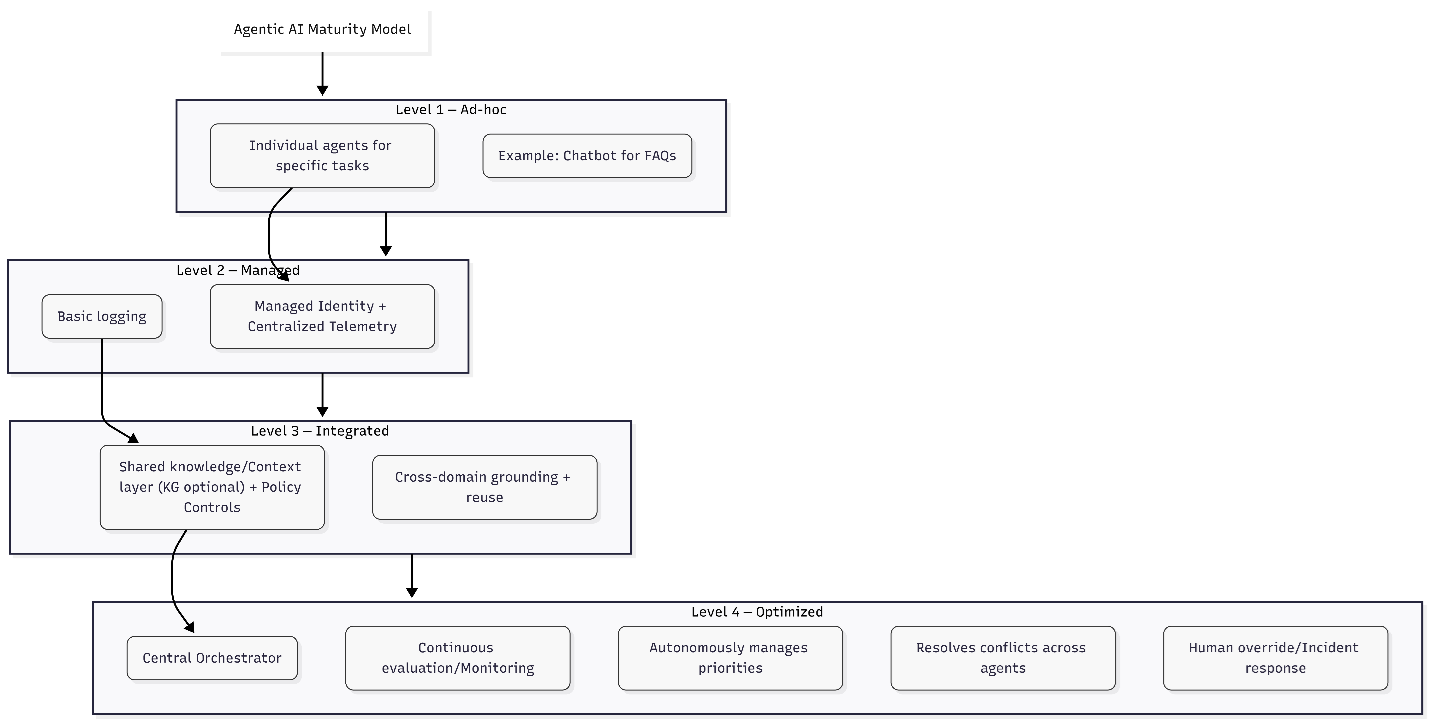}
    \caption{Visualization of Agentic AI Maturity Model}
    \label{fig:Picture1}
\end{figure}

The Agentic AI maturity model for healthcare can be framed as a progressive evolution from isolated tools to coordinated, autonomous multi‑agent ecosystems, aligning with established AI and digital health maturity concepts in the literature \cite{Mulder2023}. Figure \ref{fig:Picture1} offers a high level view of the Agentic AI maturity model. At Level 1 (Ad‑hoc), organizations deploy single-purpose agents, such as FAQ chatbots or task-specific assistants, that operate in isolation, echoing early, narrow AI point solutions with limited integration into clinical workflows \cite{Mulder2023, Zohaib2025}. Level 2 (Managed) introduces basic operational governance through shared access controls using managed identity and logging, consistent with security and auditability requirements emphasized in healthcare AI governance and regulatory guidance \cite{Mulder2023}. As systems advance to Level 3 (Integrated), agents begin to interoperate over shared knowledge structures and context, including policy control, reflecting evidence that coordinated, data-driven decision support improves care quality, continuity, and system efficiency \cite{Mulder2023}. At Level 4 (Optimized), a central orchestrator dynamically prioritizes tasks, coordinates agents, and resolves conflicts across workflows, aligning with emerging descriptions of Agentic AI as an autonomy layer that manages multi-step plans, orchestrates tools, and adapts to complex healthcare environments while remaining subject to oversight, safety constraints, and governance\cite{Mulder2023, Zohaib2025}.

\subsection{An Adoption-Centered Governance Framework}
\begin{figure}[!h]
    \centering
    \includegraphics[width=0.70\linewidth]{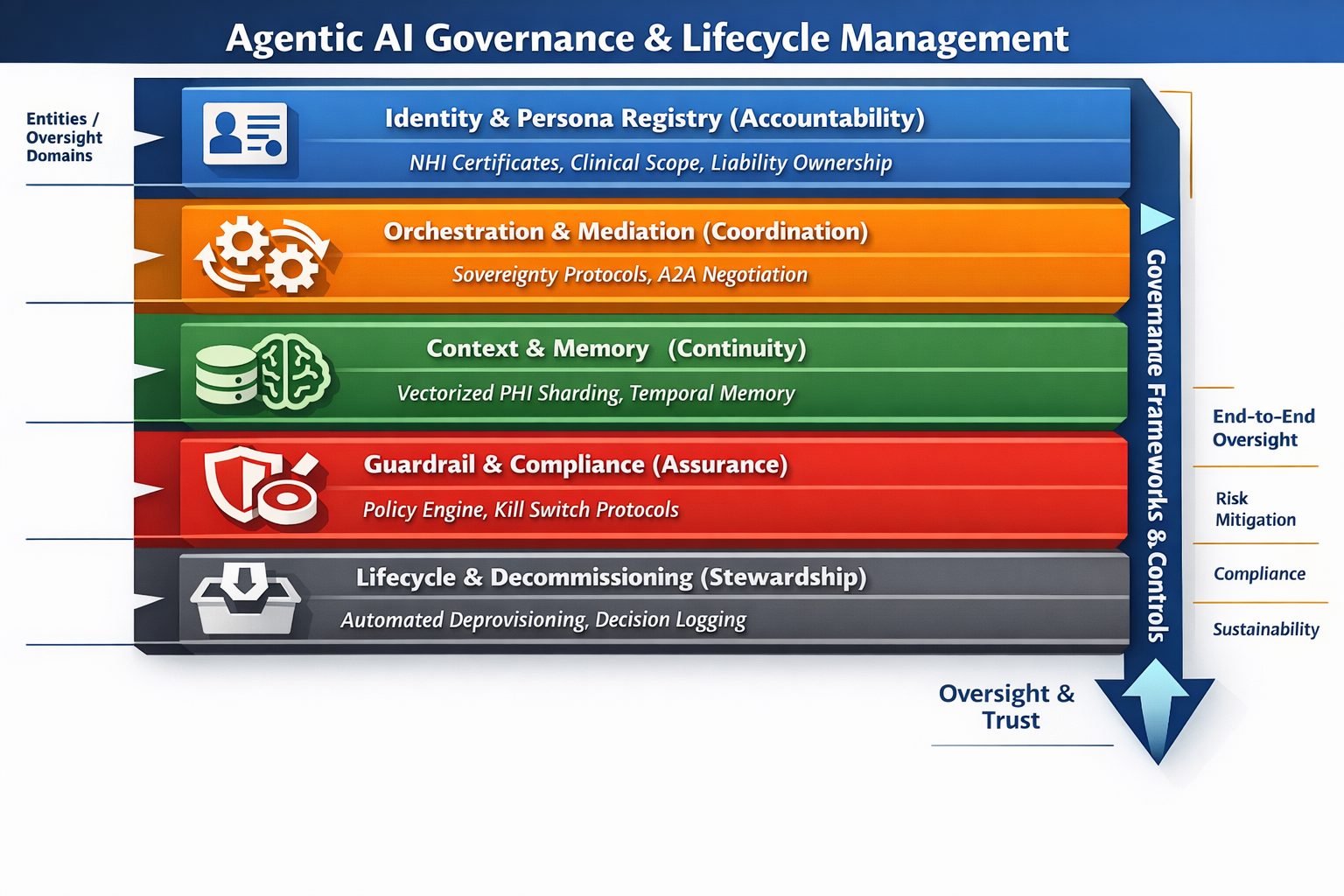}
    \caption{Agentic AI Governance and Lifecycle Management Framework}
    \label{fig:AgenticAILifecycleOverview}
\end{figure}

We propose a five-layer framework to govern Agentic AI fleets and reduce agent sprawl in healthcare. Figures \ref{fig:AgenticAILifecycleOverview} and \ref{fig:Picture3} present a system architecture diagram that separates the architecture into five layers of responsibility to manage and govern the agent lifecycle. 
\begin{itemize}
    \item \textbf{Layer 1-} Identity \& Persona Registry (Accountability): Layer 1 serves as a single system of record for every agent in the enterprise, representing ownership, semantic capabilities, responsibility, and traceability for each agent. The centralized agent registry is intended to address agent redundancy and must also enforce the principle of least privilege. The key components involved at this layer are NHI (Non-Human Identity) certificates, Clinical Scope of Practice definitions, and liability ownership, which identifies the accountable individual or unit responsible for managing this agent.
    \item \textbf{Layer 2-} Orchestration \& Mediation Layer (Coordination): Layer 2 serves as an orchestration and mediation layer, managing communication and resolving conflicts between agents. It captures structured interactions and translates semantic intent across domains, focusing on policy precedence and ownership rules. This layer ensures that domain objectives and constraints prevail during inter-agent operations. This layer integrates policy-based access control, goal prioritization, authority mapping, and risk containment, prioritizing clinical outcomes over administrative costs. For example, a discharge-planning agent cannot override a medication-safety agent when clinical risk is unresolved during conflicts and functions as an agent-to-agent negotiation engine. 
    \item \textbf{Layer 3-} Context\& Memory Layer (Continuity): To ensure HIPAA compliance, Layer 3 provides long-term continuity without compromising sensitive data. Key components include PHI segmentation for retrieval, vector-store access controls (Vectorized PHI Sharding), and retention-bound longitudinal context (Temporal Memory). This setup is designed to reduce the likelihood that agents access only the necessary patient information, adhering to the principle of least privilege. Retention-bound context records patient history, supporting continuity of care.  
    \item \textbf{Layer 4-} Guardrail \& Compliance Layer (Assurance): Layer 4 provides real-time monitoring and kill-switch protocols for active oversight and risk containment. It includes supervisor agents that monitor other agents and use Governance-as-Code (GAC) to ensure every action is verified against a policy engine. This can block or escalate unauthorized clinical actions, such as drug changes without human oversight. 
    \item \textbf{Layer 5-} Lifecycle \& Decommissioning Layer (Stewardship): Layer 5 governs provisioning, change control, retirement, and evidence retention, and is directly responsible for agent lifecycle management, from agent provisioning to decommissioning. Every active agent must have a defined expiration date, reflecting end-to-end custodianship. This layer addresses concerns related to agent drift and task completion, automatically revoking all NHI tokens and maintaining agent decision logs. 
\end{itemize}

\subsection{KPIs and Metrics}
Leaders need operational tools to manage agent fleets, much as they manage clinical systems, including clear accountability, disciplined change control, auditability, and measurable safety. The framework proposes a small KPI set aligned with the enterprise AI governance council to support routine oversight and early course-correction, before isolated pilots quietly turn into enterprise-wide control gaps:
\begin{itemize}
    \item	\% of agents with a named accountable owner recorded in the registry
    \item	Median time to revoke agent credentials after retirement or scope change
    \item	\% of tool calls with a recorded policy decision (allow/deny) and policy version
    \item	Orphan-agent count: agents running without active ownership or approval
    \item	PHI-minimization rate: proportion of workflows limited to the minimum necessary data via mediated interfaces
    \item	Control drift rate: \% of agents operating outside the approved baseline (policy/model/prompt/config)
    \item	Agent-related incident rate: events tied to agent behavior (e.g., tool misuse, unintended PHI exposure)

\end{itemize}

For simulation-based plausibility analysis, we refine each level with measurable KPI thresholds as presented in Table \ref{tab:maturitymodel}. Advancement requires all seven KPIs to meet the target level’s threshold simultaneously (strict rule), ensuring that level assignment is confirmatory rather than subjective.

\begin{longtable}{p{3.5cm}p{2.0cm}p{2.0cm}p{2.0cm}p{2.0cm}p{3.5cm}}
    \caption{Maturity Model with Measurable KPI Thresholds.}
    \label{tab:maturitymodel} \\
        \toprule
        \textbf{KPI} & \textbf{Level 1 (Ad-hoc)} & \textbf{Level 2 (Managed)} & \textbf{Level 3 (Integrated)} & \textbf{Level 4 (Optimized)}  & \textbf{Rationale/source} \\ 
        \midrule
        \endfirsthead
        \multicolumn{5}{c}{\small \textit{Table \ref{tab:maturitymodel} (continued)}}\\
        \toprule
        \textbf{KPI} & \textbf{Level 1 (Ad-hoc)} & \textbf{Level 2 (Managed)} & \textbf{Level 3 (Integrated)} & \textbf{Level 4 (Optimized)} & \textbf{Rationale/source} \\  
        \midrule
        \endhead
        
        \midrule
        \multicolumn{5}{r}{\small \textit{Table continues on next page}}\\
        \endfoot
        
        \bottomrule
        \endlastfoot
        
        \% Agents with Named Owner & <50\% & $\geq$80\% & $\geq$95\% & $\geq$99\% & Expert-Derived \\
        \midrule
        Median Credential Revocation  &	>30 days & 	$\leq$14 days &	$\leq$7 days & $\leq$48 hours & Expert-Derived\\
        \midrule
        \% Tool Calls Logged &	<25\% &	$\geq$60\% &	$\geq$85\% &	$\geq$95\% & Regulatory expectation\\
        \midrule
        Orphan-Agent Ratio &	>20\% &	$\leq$10\% &	$\leq$5\% &	$\leq$2\% & Simulation parameter\\
        \midrule
        PHI-Minimization Rate &	<40\% &	$\geq$60\% &	$\geq$80\% &	$\geq$95\% & Regulatory expectation\\
        \midrule 
        Control Drift Rate &	>30\% &	$\leq$15\% &	$\leq$8\% &	$\leq$3\% & Assumption\\
        \midrule
        Incident Rate (per agent/month) &	>0.15 &	$\leq$0.10 &	$\leq$0.05 &	$\leq$0.02 & Expert-Derived\\
                
\end{longtable}

\section{Simulation-Based Evaluation Method}
To evaluate the expected behavior of UALM under controlled assumptions, we designed a Monte Carlo simulation with stochastic event generation. This approach was chosen because real-world deployment data from UALM-governed agent fleets does not yet exist, and simulation allows controlled comparison across governance conditions while maintaining reproducibility \cite{rubinstein2017simulation}. Therefore, results should be interpreted as model-based projections rather than observed performance. \\

\subsection{Simulation Design Overview}
The simulation models agent fleet operations over a six-month pilot phase across three synthetic healthcare organizations of varying size and complexity. Each organization is simulated under four governance conditions, with 1,000 independent Monte Carlo runs per organization-condition pair (totaling 12,000 runs for the main evaluation). Each run tracks seven KPIs monthly, with stochastic events injected based on agent-level probability distributions. We selected 1,000 Monte Carlo replications per organization-condition pair to reduce sampling variability and provide stable estimates of KPI means and confidence intervals. Because the simulation compares four governance conditions across three organization sizes, a balanced replication design was used to prevent differences in statistical precision across conditions. Prior to final submission, convergence should be confirmed by reporting cumulative KPI means at 250, 500, 750, and 1,000 runs. This check will demonstrate whether additional replications materially change the estimated KPI values or the rank ordering of governance conditions. \\

A six-month simulation horizon was selected to represent an initial governance implementation window rather than a mature enterprise steady state. In healthcare organizations, agentic AI governance typically begins as a controlled pilot before broader institutionalization; therefore, six-monthly time steps provide sufficient time to observe early agent growth, governance ramp-up, control drift, orphan-agent formation, credential revocation behavior, and incident accumulation. The six-month period also allows comparison of governance trajectories before the model becomes dominated by long-term organizational assumptions that cannot yet be empirically calibrated. Accordingly, the simulation should be interpreted as an early-stage adoption analysis rather than a long-term forecast of agent governance performance.

\subsection{Synthetic Organization Profiles}
Three healthcare organizations were constructed to represent the spectrum of health system sizes as presented in \ref{tab:syntheticprofiles}. Agent types were drawn from practical healthcare use cases: Clinical Documentation Assistant, Sepsis Early Warning Monitor, Insurance Eligibility Checker, Patient Scheduling Bot, Revenue Cycle Coding Assistant, Patient Portal Chatbot, Medication Reconciliation Agent, Radiology Triage Assistant, Prior Authorization Agent, Lab Result Notification Agent, Discharge Planning Assistant, and Bed Management Optimizer. Each agent type was assigned a clinical criticality score (0.2–0.95) and a PHI exposure level (0.3–0.9), both of which influenced event probabilities.

\begin{longtable}{p{2.0cm}p{2.0cm}p{3.0cm}p{2.0cm}}
    \caption{Synthetic Organization Profiles.}
    \label{tab:syntheticprofiles} \\
        \toprule
        \textbf{Parameter} & \textbf{Small} & \textbf{Medium} & \textbf{Large}  \\ 
        \midrule
        \endfirsthead
        \multicolumn{4}{c}{\small \textit{Table \ref{tab:syntheticprofiles} (continued)}}\\
        \toprule
       \textbf{Parameter} & \textbf{Small} & \textbf{Medium} & \textbf{Large}  \\ 
        \midrule
        \endhead
        
        \midrule
        \multicolumn{4}{r}{\small \textit{Table continues on next page}}\\
        \endfoot
        
        \bottomrule
        \endlastfoot
        
        Type & 	Community Hospital & 		Regional System & 	Academic Medical Center \\
         \midrule
        Bed Count & 	150 & 	400 & 	800 \\
         \midrule
        Starting Agent Fleet & 	4 agents & 	10 agents & 	22 agents \\
         \midrule
        Monthly Growth Rate & 	$\lambda$= 1.0 & 	$\lambda$= 2.0 & 	$\lambda$ = 3.5 \\
         \midrule
        Available Agent Types & 	6 types & 	10 types & 	12 types \\
                
\end{longtable}

The three synthetic organizations were designed as stylized profiles rather than population estimates. The 150-bed profile represents a community hospital with a limited number of agentic AI deployments concentrated in documentation, scheduling, eligibility, and patient communication. The 400-bed profile represents a regional health system with broader clinical, administrative, and revenue-cycle automation, requiring more agent types and higher monthly growth. The 800-bed profile represents an academic medical center or large integrated delivery environment where agentic AI use cases are more likely to span clinical monitoring, specialty triage, discharge planning, bed management, revenue cycle, and patient access. Starting fleet sizes of 4, 10, and 22 agents were selected to reflect increasing operational complexity across these profiles while keeping the model focused on early fleet growth rather than highly mature AI ecosystems. These values should be interpreted as scenario assumptions used for comparative analysis, not as empirical estimates of current healthcare deployment levels.

\subsection{Governance Conditions}
Four governance conditions were compared to evaluate both the absolute and incremental value of UALM as presented in the Table \ref{tab:govcondition}. The Registry-Only condition was included to test whether a simple agent inventory (which many organizations already maintain) provides sufficient governance. The NIST RMF-Lite condition represents the application of generic AI risk management principles \cite{NIST2024} without agent-specific lifecycle controls, testing whether UALM’s agent-specific architecture adds value beyond established frameworks.
\begin{longtable}{p{3.0cm}p{4.0cm}p{7.0cm}}
    \caption{Governance Conditions Compared in the Simulation.}
    \label{tab:govcondition} \\
        \toprule
        \textbf{Condition} & \textbf{Description} & \textbf{Capabilities}  \\ 
        \midrule
        \endfirsthead
        \multicolumn{3}{c}{\small \textit{Table \ref{tab:govcondition} (continued)}}\\
        \toprule
       \textbf{Condition} & \textbf{Description} & \textbf{Capabilities}  \\ 
        \midrule
        \endhead
        
        \midrule
        \multicolumn{3}{r}{\small \textit{Table continues on next page}}\\
        \endfoot
        
        \bottomrule
        \endlastfoot
        
        No Governance &   	Status quo: agents deployed without centralized oversight & No detection or mitigation; ownership and logging degrade naturally over time \\
         \midrule
        Registry-Only  & 	Layer 1 only: centralized agent inventory with identity management & 	Strong duplicate detection (85\%), moderate orphan tracking (70\%), minimal coverage for drift, PHI, or injection events \\
         \midrule
        NIST RMF-Lite & 	Generic AI risk management framework applied to agents (not agent-specific) & 	Moderate detection (25–50\%) across all event types, moderate logging and PHI controls, no agent-specific lifecycle automation \\
         \midrule
        UALM (Full)  & 	All five UALM layers operational & 	Full detection capabilities (78–92\%), policy-as-code enforcement, automated lifecycle management, PHI segmentation, orchestration \\

\end{longtable}

\subsection{Stochastic Event Generation}
Seven event types were injected stochastically each simulation month. The event categories were selected to represent plausible governance failure modes in healthcare agent fleet operations, including drift, orphaned ownership, excess PHI access, duplicate deployment, vendor deprecation, inter-agent conflict, and prompt injection. Because no benchmark dataset currently exists for UALM-governed healthcare agent fleets, the base probabilities were treated as transparent scenario parameters rather than empirically calibrated rates (Table \ref{tab:StochasticEventTypes}). Event probabilities were adjusted per-agent based on clinical criticality and PHI exposure levels. For PHI near-misses and agent drift, probabilities are scaled as (0.7 + 0.6 × PHI\_exposure); for prompt injection, as (0.8 + 0.4 × criticality). This ensured that high-risk agents (e.g., Sepsis Early Warning Monitor, Medication Reconciliation Agent) experienced a proportionally greater number of events.
\begin{longtable}{p{3.0cm}p{2.5cm}p{3.5cm}p{6.0cm}}
    \caption{Stochastic Event Types and Parameters}
    \label{tab:StochasticEventTypes} \\
        \toprule
        \textbf{Event Type} & \textbf{Base Probability} & \textbf{Responsible UALM Layer} & \textbf{Clinical Relevance}  \\ 
        \midrule
        \endfirsthead
        \multicolumn{4}{c}{\small \textit{Table \ref{tab:StochasticEventTypes} (continued)}}\\
        \toprule
        \textbf{Event Type} & \textbf{Base Probability} & \textbf{Responsible UALM Layer} & \textbf{Clinical Relevance}  \\ 
        \midrule
        \endhead
        
        \midrule
        \multicolumn{4}{r}{\small \textit{Table continues on next page}}\\
        \endfoot
        
        \bottomrule
        \endlastfoot

        Agent Drift & 	0.08/agent/month	 & Layer 4 (Guardrails) & 	Agent deviates from approved clinical scope \\
         \midrule
        Orphan Event & 	0.05/agent/month & 	Layer 1 (Registry) & 	Owner departs or is reassigned; agent persists \\
         \midrule
        PHI Near-Miss & 	0.06/agent/month & 	Layer 3 (Context) & 	Agent accesses excess protected health information \\
         \midrule
        Duplicate Deployment & 	0.04/agent/month & 	Layer 1 (Registry) & 	New agent overlaps existing capability \\
         \midrule
        Vendor Deprecation & 	0.02/agent/month & 	Layer 5 (Lifecycle)	 & Vendor drops support; agent becomes unsupported \\
         \midrule
        Inter-Agent Conflict & 	0.03/agent/month & 	Layer 2 (Orchestration) & 	Two agents produce conflicting clinical outputs \\
         \midrule
        Prompt Injection	 & 0.04/agent/month & 	Layer 4  & (Guardrails)	Adversarial input attempt on agent \\

\end{longtable}
The event probabilities were intentionally specified at modest monthly rates to model low-frequency but operationally meaningful governance failures. Higher-risk agents were assigned proportionally greater exposure through PHI and clinical criticality multipliers, reflecting the assumption that agents with broader access to sensitive data or clinical workflow influence create larger governance surfaces. These assumptions were subsequently varied in sensitivity analysis to test whether the comparative findings depended on a narrow parameter setting.

\subsection{Governance Ramp-Up Curve}
All governance conditions incorporated a temporal ramp-up curve reflecting real-world implementation dynamics. The UALM default curve progressed from 50\% effectiveness at month 1 to 97\% at month 6: [0.50, 0.65, 0.78, 0.87, 0.93, 0.97]. Registry-Only and NIST RMF-Lite had analogous but lower curves. This design choice acknowledges that governance effectiveness improves over time as processes mature, and was itself subjected to sensitivity analysis.

\subsection{UALM Single-Point-of-Failure Modeling}
The 0.5\% monthly outage probability was selected as a low-frequency but high-impact stressor to represent partial degradation of a centralized governance control plane. This assumption was not intended to estimate a specific vendor or infrastructure failure rate. Rather, it was included to test whether UALM’s benefits remain credible when its central coordination layer experiences intermittent impairment. During such events, detection effectiveness was reduced by 70\% to represent degraded policy enforcement, logging, and orchestration capacity while assuming that some local controls remain operational.

To test whether the results were sensitive to this assumption, the outage probability should be varied across four levels: 0.1\%, 0.5\%, 1.0\%, and 2.0\% per month. The 0.\% condition represents highly resilient control-plane operations; 0.5\% represents the default low-frequency disruption assumption; 1.0\% represents moderate reliability degradation; and 2.\% represents a stressed centralization scenario. Reporting UALM’s incident-rate reduction across these outage levels would make the single-point-of-failure analysis more transparent and prevent the centralization risk from appearing understated.

\subsection{Statistical Analysis}
Statistical comparisons used non-parametric Mann-Whitney U tests (appropriate given the non-normal distribution of simulation outputs) with Cohen’s d effect sizes and 95\% confidence intervals computed from t-distributions. Five pairwise comparisons were conducted for each KPI across each organization: Registry-Only vs. No Governance, NIST RMF-Lite vs. No Governance, UALM vs. No Governance, UALM vs. Registry-Only, and UALM vs. NIST RMF-Lite (totaling 105 comparisons: 5 pairs × 7 KPIs × 3 organizations).

\subsection{Sensitivity Analyses}
Two sensitivity analyses were conducted. First, a parameter sensitivity analysis varied event probabilities (0.5×, 1.5×) and agent growth rates (0.5×, 2.0×) across six scenarios, including a worst-case combination (1.5× events, 2.0× growth), with 500 runs per condition. Second, a governance ramp-up sensitivity analysis tested four adoption curves for UALM: slow adoption [0.25–0.72], default [0.50–0.97], aggressive [0.70–0.99], and stalled-at-mid [0.50–0.74].

\subsection{Reproducibility}
All simulation code was implemented in Python using NumPy for stochastic generation, Pandas for data management, SciPy for statistical tests, and Matplotlib/Seaborn for visualization. The random seed was fixed at 42 for full reproducibility. The complete codebase (ualm\_simulation.py and ualm\_analysis.py) is provided as supplementary material.

\section{Results}

\subsection{Comparative Advantage: UALM vs. Governance Alternatives}
\begin{figure}[!h]
    \centering
    \includegraphics[width=1.0\linewidth]{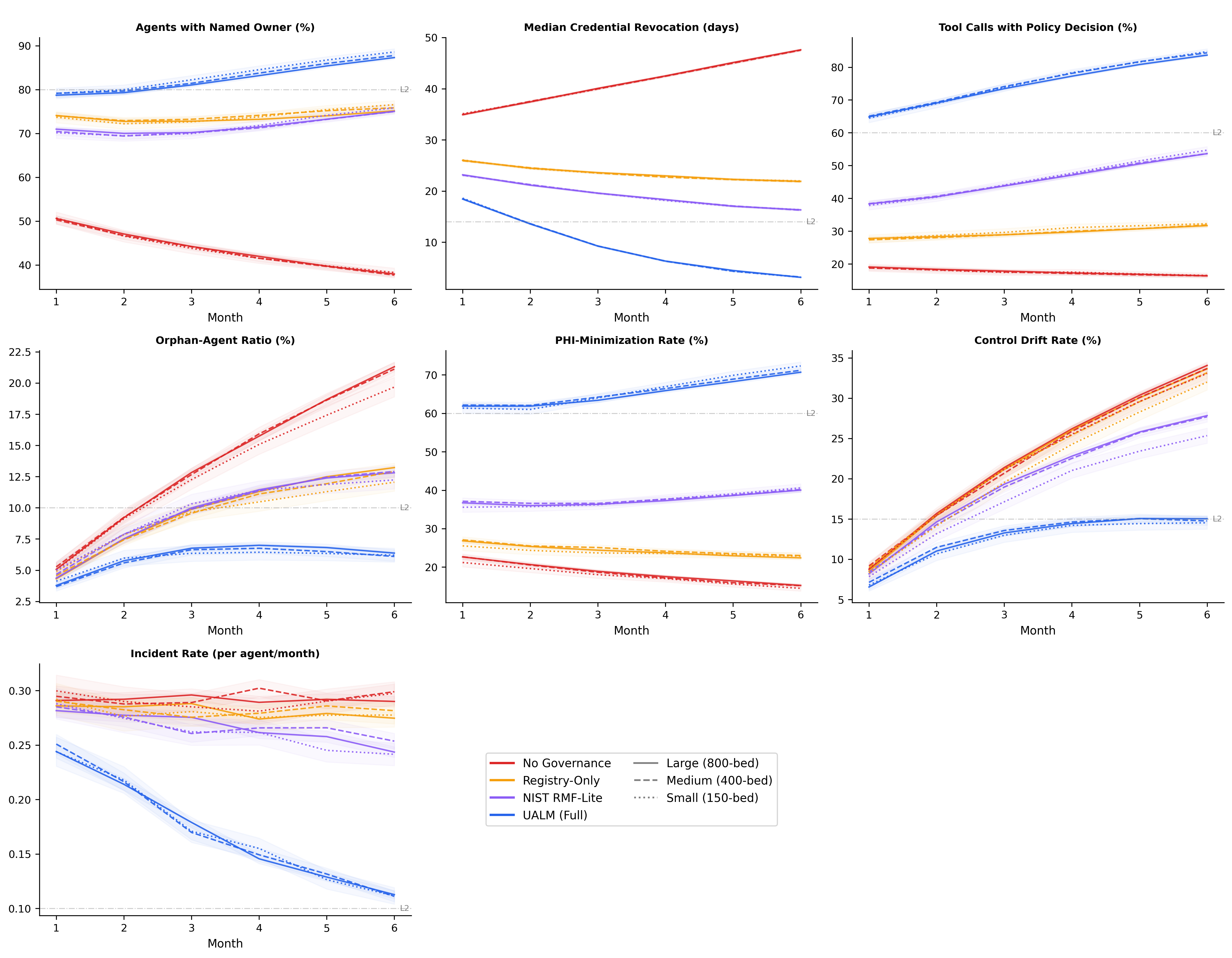}
    \caption{KPI Trajectories Across Four Governance Conditions (Mean ± 95\% CI, N = 1,000 runs per condition)}
    \label{fig:KPITrajectories}
\end{figure}

Figure \ref{fig:KPITrajectories} summarizes KPI trajectories across the four governance conditions over the six-month simulation horizon. The results are presented primarily in terms of practical effect size and directional consistency rather than statistical significance alone, because the simulation uses repeated stochastic replications from specified model assumptions. Across all three organization profiles, UALM produced the most favorable month-six outcomes on all seven KPIs, with the largest gains observed in credential revocation time, tool-call logging, PHI minimization, and incident reduction. The most informative comparison is UALM versus NIST RMF-Lite, because this contrast tests whether agent-specific lifecycle governance adds value beyond a generic AI risk management approach. Statistical tests were retained as confirmatory checks, but interpretation focuses on the magnitude, stability, and operational relevance of the observed differences.

The most informative comparison is UALM vs. NIST RMF-Lite, as it tests whether agent-specific lifecycle governance adds value beyond generic AI risk management. Table \ref{tab:UALMvsNIST} presents the key results for the medium-sized organization at month 6.

\begin{longtable}{p{5.0cm}p{2.5cm}p{2.5cm}p{2.5cm}p{2.5cm}}
    \caption{UALM vs. NIST RMF-Lite: Medium Organization at Month 6 (*** p < 0.001)}
    \label{tab:UALMvsNIST} \\
        \toprule
        \textbf{KPI} & \textbf{UALM} & \textbf{NIST RMF-Lite} & \textbf{Improvement} & \textbf{Cohen’s d} \\ 
        \midrule
        \endfirsthead
        \multicolumn{5}{c}{\small \textit{Table \ref{tab:UALMvsNIST} (continued)}}\\
        \toprule
        \textbf{KPI} & \textbf{UALM} & \textbf{NIST RMF-Lite} & \textbf{Improvement} & \textbf{Cohen’s d} \\ 
        \midrule
        \endhead
        
        \midrule
        \multicolumn{5}{r}{\small \textit{Table continues on next page}}\\
        \endfoot
        
        \bottomrule
        \endlastfoot
        Agents with Owner (\%) & 		89.4 & 		76.5 & 		+17\% & 		1.42*** \\
         \midrule
        Credential Revocation (days) & 		3.1 & 		15.5 & 		-80\% & 		6.06*** \\
         \midrule
        Tool Calls Logged (\%) & 		86.5 & 		54.9 & 		+57\% & 		2.95*** \\
         \midrule
        Orphan-Agent Ratio (\%) & 		6.4	 & 	12.4 & 		-53\% & 		0.99*** \\
         \midrule
        PHI-Minimization Rate (\%) & 		71.7	 & 	40.5 & 		+77\% & 		2.83*** \\
         \midrule
        Control Drift Rate (\%) & 		14.6 & 		27.3 & 		-47\% & 		1.34*** \\
         \midrule
        Incident Rate (per agent/month) & 		0.12 & 		0.24 & 		-56\% & 		1.45*** \\
\end{longtable}

\begin{figure}[!h]
    \centering
    \includegraphics[width=1\linewidth]{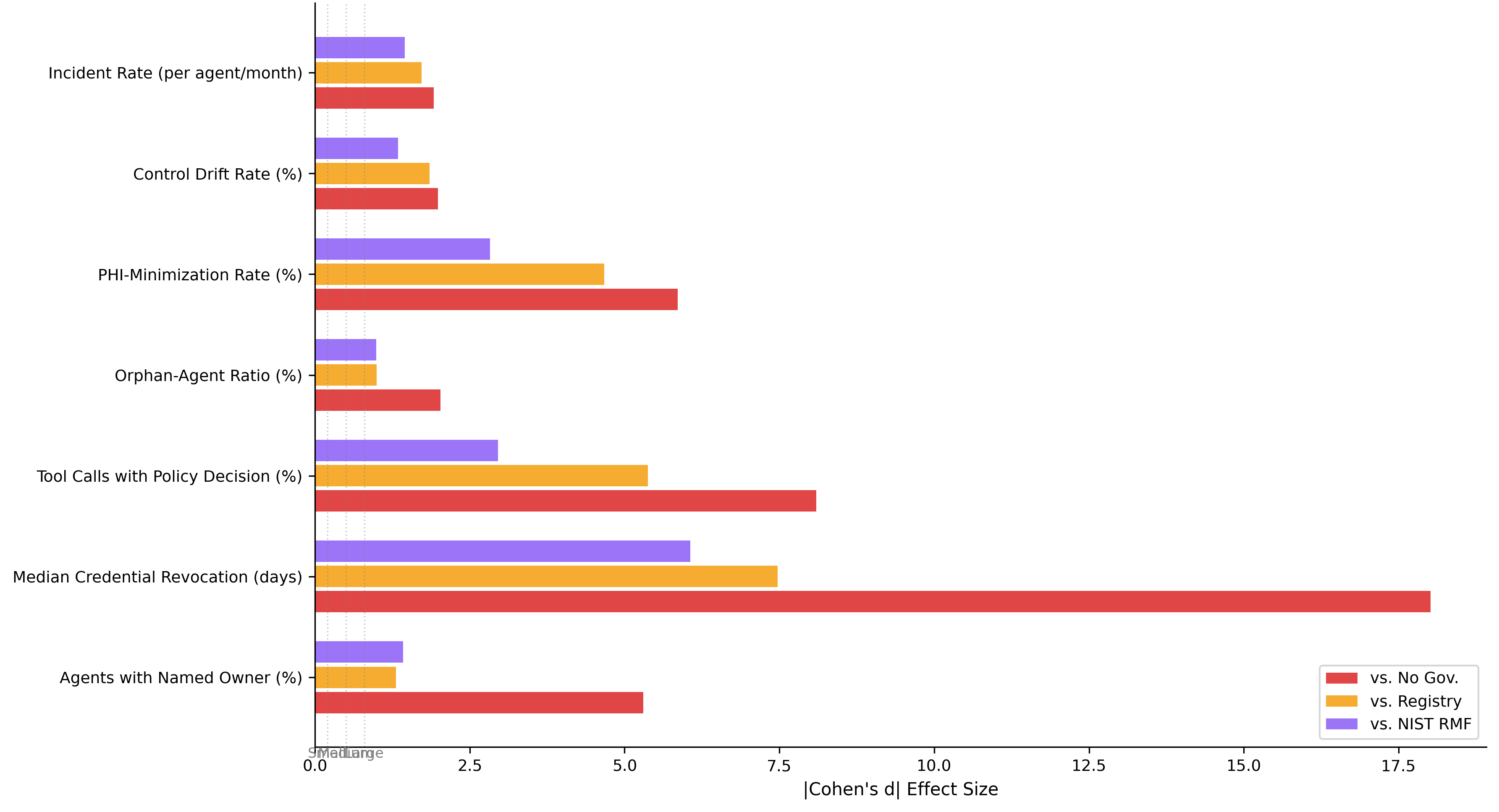}
    \caption{Effect Sizes: UALM vs. Each Comparator (Medium Organization, Month 6; all p < 0.001)}
    \label{fig:fig7_effect_sizes}
\end{figure}

UALM’s largest advantages were in credential revocation time (d = 6.06), reflecting the automated lifecycle management of Layer 5, and in tool-call logging (d = 2.95), reflecting Layer 4’s Governance-as-Code enforcement. Against the Registry-Only condition, UALM showed even larger advantages in PHI minimization (+209\%, d = 4.67) and tool-call logging (+165\%, d = 5.38), confirming that going beyond agent inventory management is essential. Figure \ref{fig:fig7_effect_sizes} presents the effect sizes for UALM relative to each comparator, illustrating the graduated improvement UALM provides over increasingly sophisticated partial-governance approaches.

\begin{figure}[!h]
    \centering
    \includegraphics[width=1\linewidth]{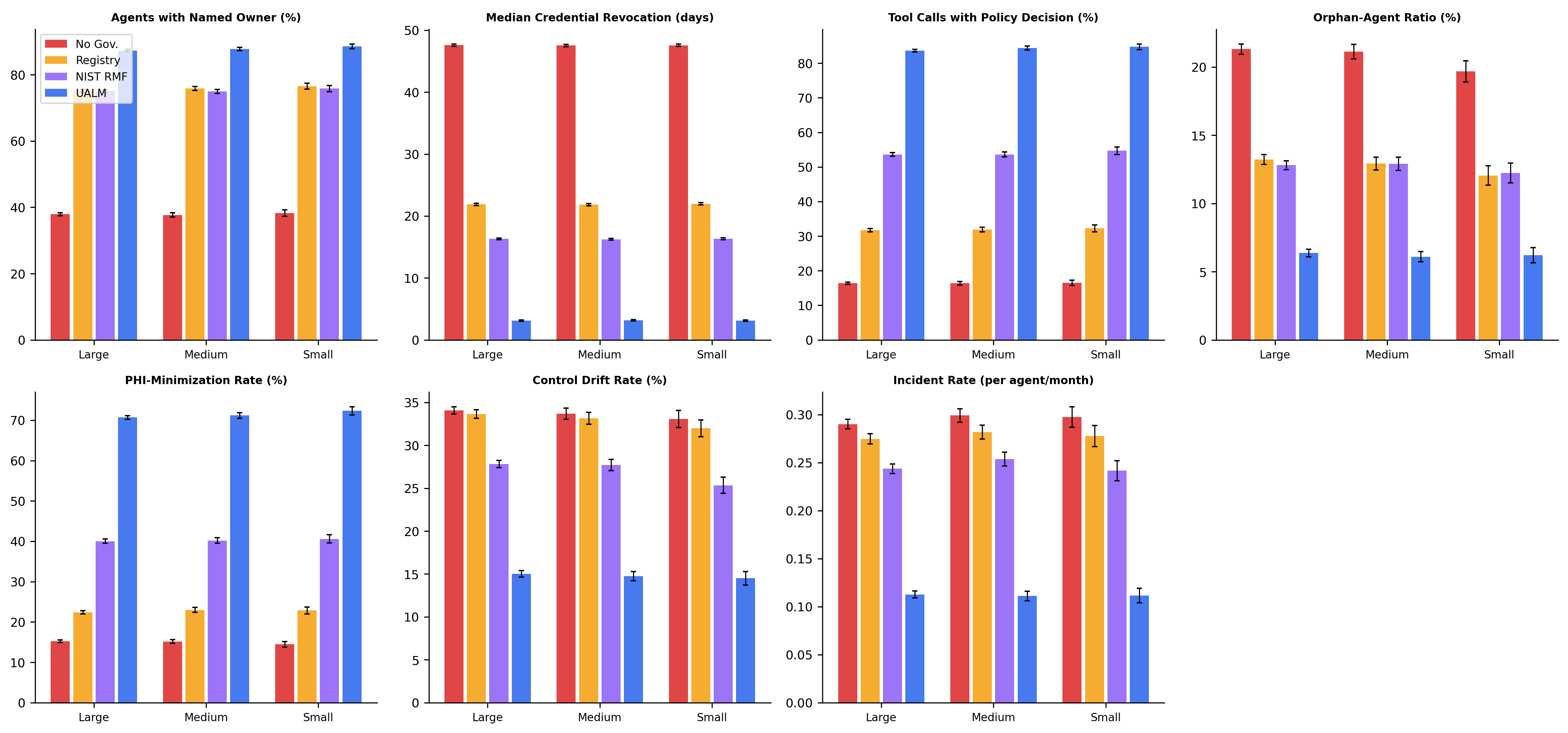}
    \caption{Month 6 KPI Comparison: All Four Governance Conditions (Mean ± 95\% CI, N = 1,000 runs)}
    \label{fig:fig12_month6_comparison}
\end{figure}

Figure \ref{fig:fig12_month6_comparison} provides a direct visual comparison of all four governance conditions across all KPIs at month 6, confirming UALM's consistent superiority across organization sizes.

\subsection{Maturity Model Validity}
\begin{figure}[!h]
    \centering
    \includegraphics[width=1\linewidth]{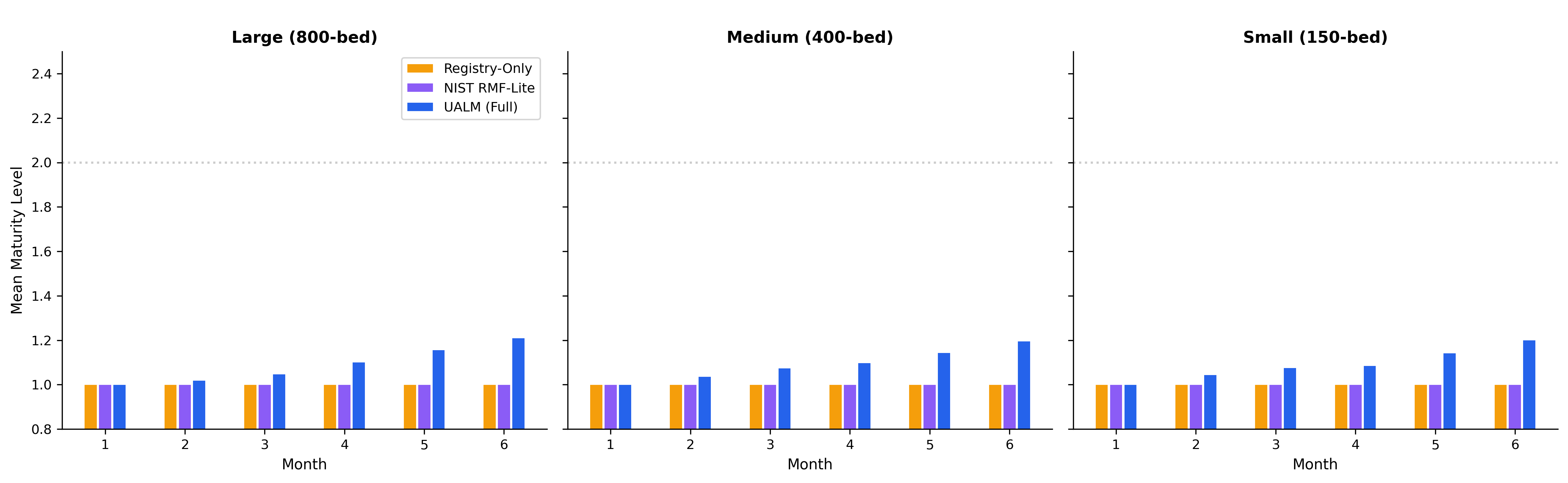}
    \caption{Mean Maturity Level Progression by Governance Condition (1,000 runs; strict all-KPI threshold rule)}
    \label{fig:fig5_maturity_progression}
\end{figure}

The refined maturity model with strict all-KPI thresholds demonstrated strong discriminant validity. The strict rule reflects the weakest-link nature of governance maturity. Figure \ref{fig:fig5_maturity_progression} shows maturity progression under each governance condition. Only UALM-governed organizations achieved Level 2 (Managed) maturity: 18.5\% of runs for the small organization, 19.5\% for the medium, and 21.1\% for the large. Neither Registry-Only nor NIST RMF-Lite reached Level 2 in any run, because both conditions structurally fail on specific KPIs (credential revocation time and tool-call logging for Registry-Only; PHI minimization and drift rate for NIST RMF-Lite). This confirms that the maturity model discriminates between governance approaches and that Level 2 represents a realistic but challenging six-month target.

\subsection{Framework Completeness: Layer Coverage Analysis}
\begin{figure}[!h]
    \centering
    \includegraphics[width=1\linewidth]{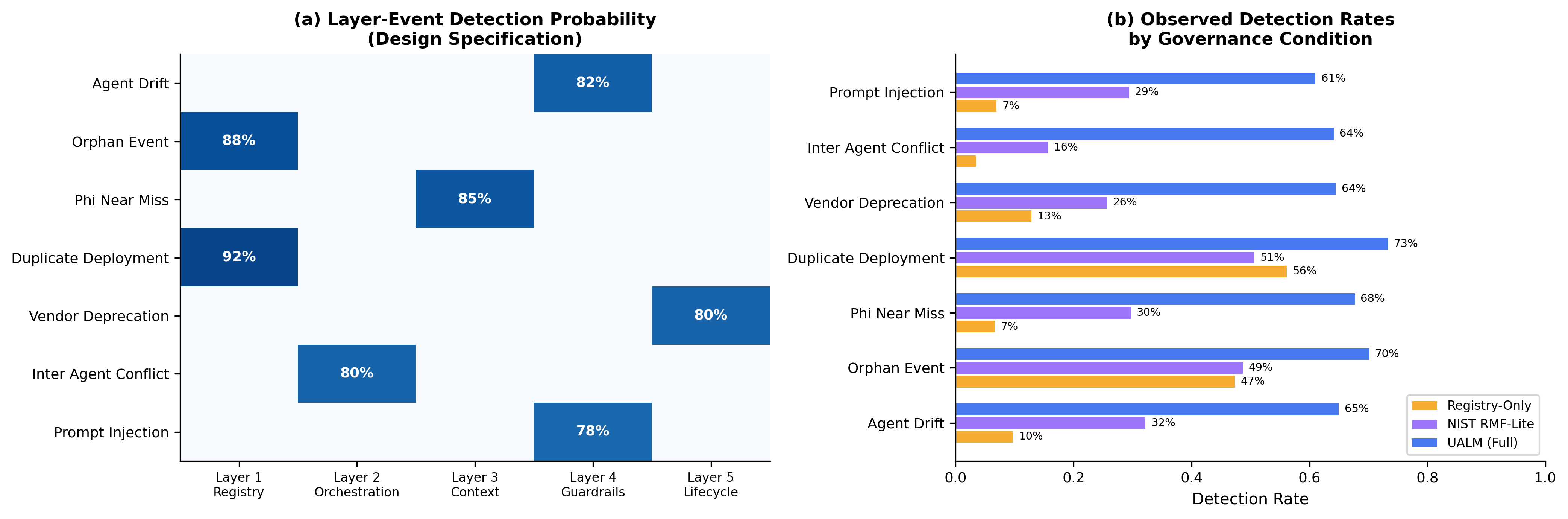}
    \caption{Framework Completeness: Layer-Event Detection Probability (Design) and Observed Detection Rates by Governance Condition}
    \label{fig:fig6_layer_coverage}
\end{figure}

Figure \ref{fig:fig6_layer_coverage} presents the framework completeness analysis, mapping each UALM layer to its intended failure modes and comparing observed detection rates across governance conditions. All seven injected failure types were mapped to and addressed by at least one UALM layer. UALM achieved detection rates of 62–74\% across event types, with the highest rates for duplicate deployment (74\%, Layer 1) and orphan events (70\%, Layer 1). Notably, UALM’s advantage was most pronounced for events requiring Layers 2–5: inter-agent conflict detection (63\% vs. 4\% for Registry-Only), PHI near-miss detection (68\% vs. 9\%), and prompt injection detection (62\% vs. 8\%). This confirms that the multi-layer architecture provides coverage that single-layer approaches cannot.

\subsection{KPI Early Warning Effectiveness}
\begin{figure}[!h]
    \centering
    \includegraphics[width=1\linewidth]{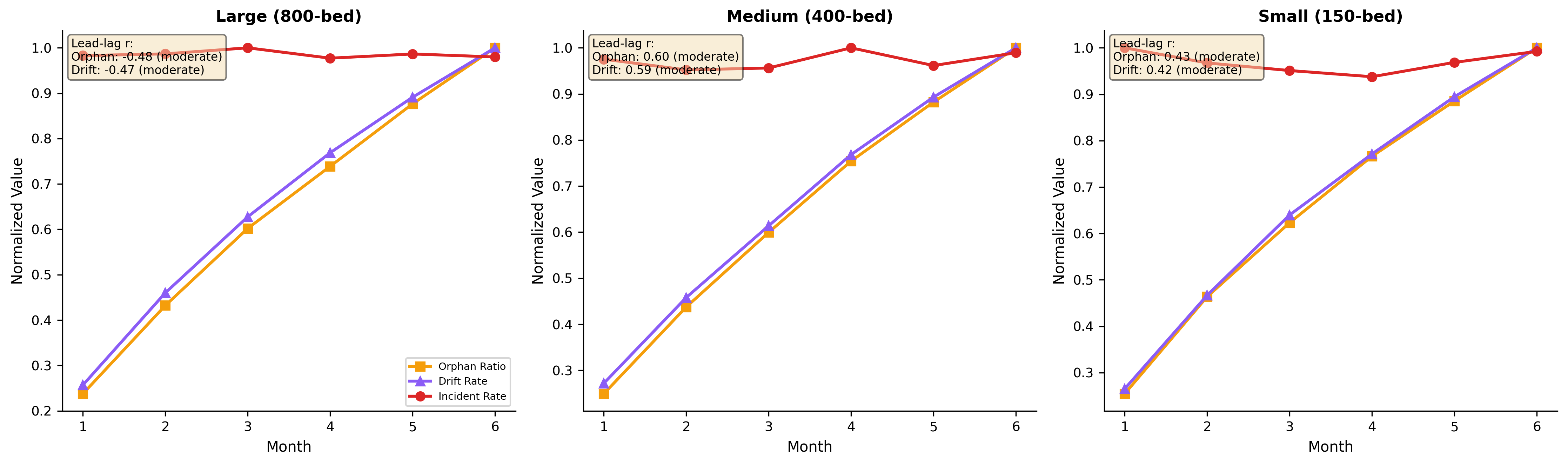}
    \caption{KPI Early Warning Effectiveness: Leading Indicators vs. Incident Rate (No-governance baseline; 1-month forecast correlation)}
    \label{fig:fig10_kpi_early_warning}
\end{figure}

We examined whether the proposed KPIs serve as leading indicators of incident escalation by computing 1-month lead-lag correlations under the no-governance baseline. Figure \ref{fig:fig10_kpi_early_warning} presents the analysis. The results were mixed. For the medium-sized organization, both orphan ratio (r = 0.60) and control drift rate (r = 0.59) showed moderate predictive value as leading indicators. For the small organization, correlations were weaker but still moderate (r $\approx$ 0.42-0.43). For the large organization, correlations were negative (r $\approx$ -0.47 to -0.48), suggesting a ceiling effect where incident rates stabilize while structural indicators continue rising. This indicates that the KPIs are stronger operational indicators than predictive ones for large-scale deployments, and fleet-size-specific thresholds may be needed for effective early warning.

\subsection{Sensitivity Analysis}
\begin{figure}[!h]
    \centering
    \includegraphics[width=1\linewidth]{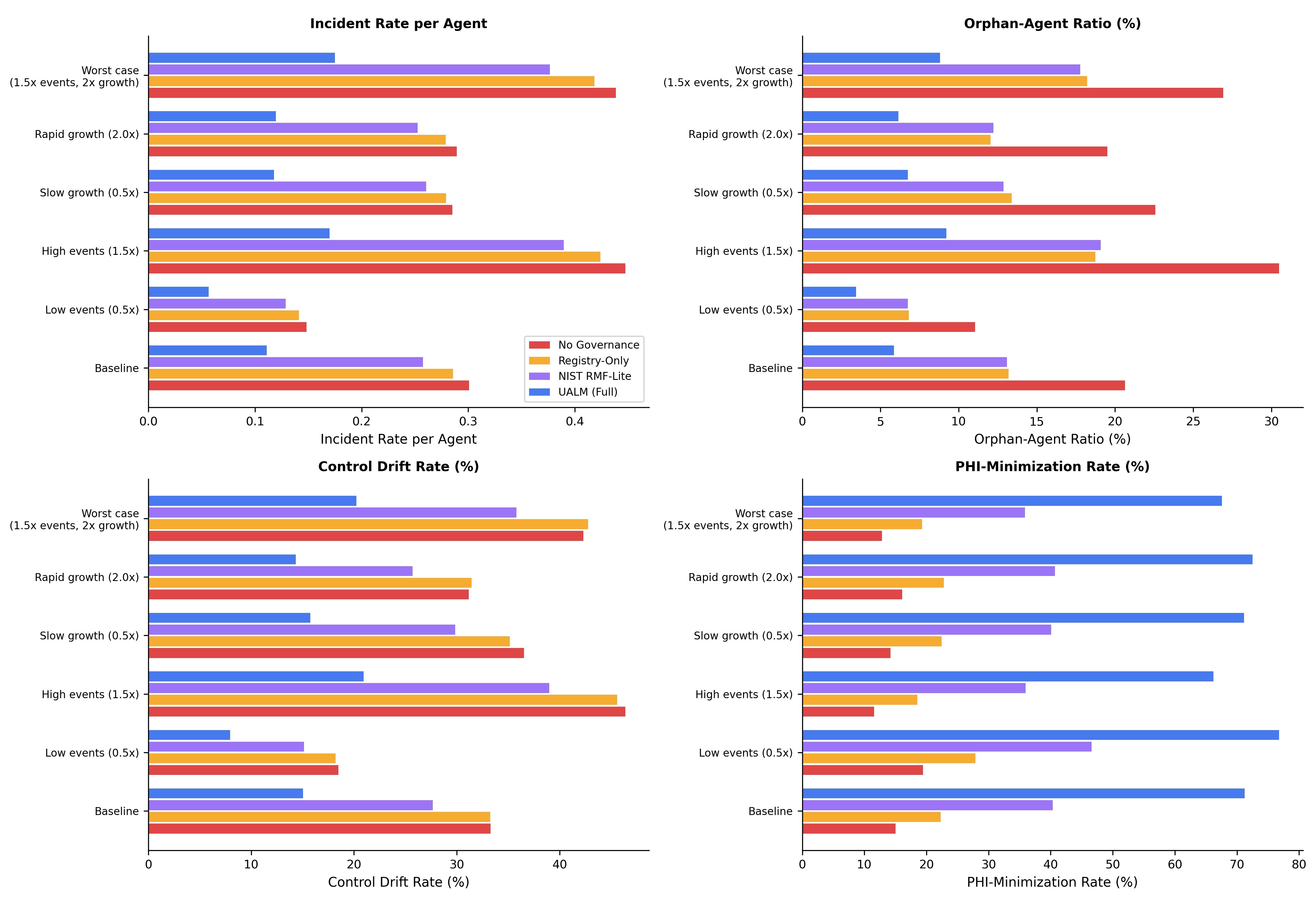}
    \caption{Sensitivity Analysis: Robustness Under Varying Conditions (Medium organization, 500 runs per scenario, Month 6)}
    \label{fig:fig8_sensitivity_analysis}
\end{figure}

UALM’s advantage was robust across all parameter variations. Figure \ref{fig:fig8_sensitivity_analysis} presents the parameter sensitivity results Even under worst-case conditions (1.5× event probability, 2.0× growth rate), UALM reduced incident rates by 56\%, only a modest decline from the 60\% baseline reduction. Rapid growth had the most impact on UALM’s relative effectiveness (57\% reduction vs. 60\% at baseline), suggesting that fleet growth management is a key operational concern.

\subsection{Governance Ramp-Up Sensitivity}
\begin{figure}[!h]
    \centering
    \includegraphics[width=1\linewidth]{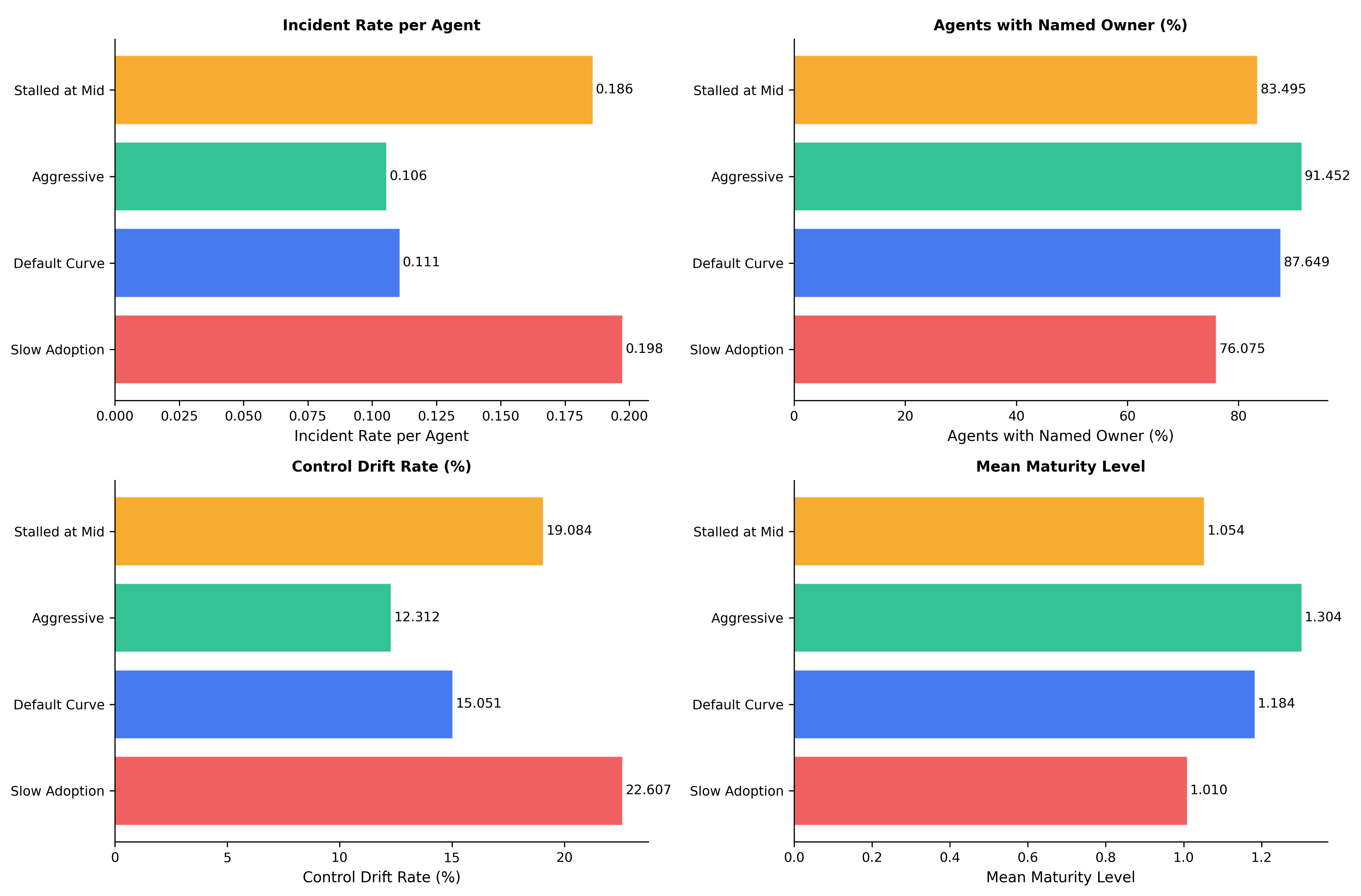}
    \caption{Governance Ramp-Up Curve Sensitivity (UALM, Medium organization, 500 runs, Month 6)}
    \label{fig:fig11_ramp_sensitivity}
\end{figure}

The assumed governance adoption curve significantly affected outcomes. Figure \ref{fig:fig11_ramp_sensitivity} presents the ramp-up sensitivity analysis. When adoption stalled at mid-level effectiveness (\textasciitilde70-74\%), incident rates remained nearly as high as slow adoption (0.186 vs. 0.198 per agent per month), compared to 0.111 under the default ramp-up curve. This is a critical finding for practitioners: UALM requires sustained organizational commitment through the full ramp-up curve. A “deploy and forget” approach will not deliver the framework’s promised benefits. Conversely, aggressive adoption (a ramp-up curve reaching 99\% by month 6) provided only a modest additional improvement over the default curve (0.106 vs. 0.111), suggesting diminishing returns beyond the default adoption trajectory.

\subsection{Quantified Drawback Analysis}
\begin{figure}[!h]
    \centering
    \includegraphics[width=1\linewidth]{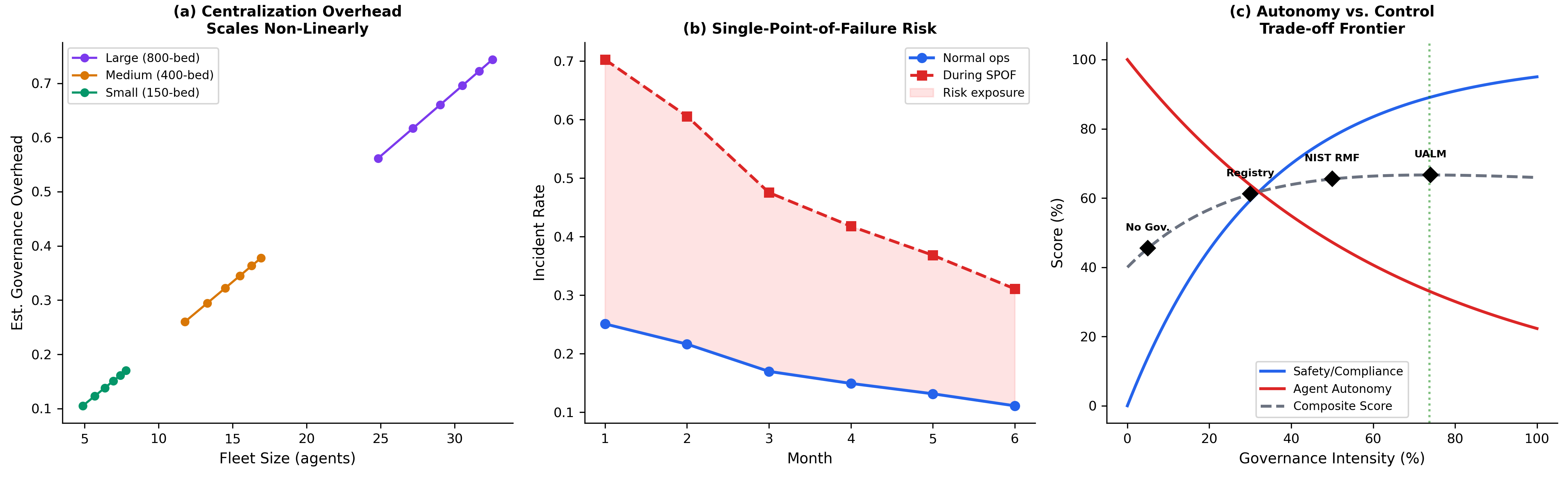}
    \caption{UALM Drawback Analysis: Centralization Overhead, Single-Point-of-Failure Risk, and Autonomy-Control Trade-off}
    \label{fig:fig9_drawback_analysis}
\end{figure}

\begin{figure}[!h]
    \centering
    \includegraphics[width=1.0\linewidth]{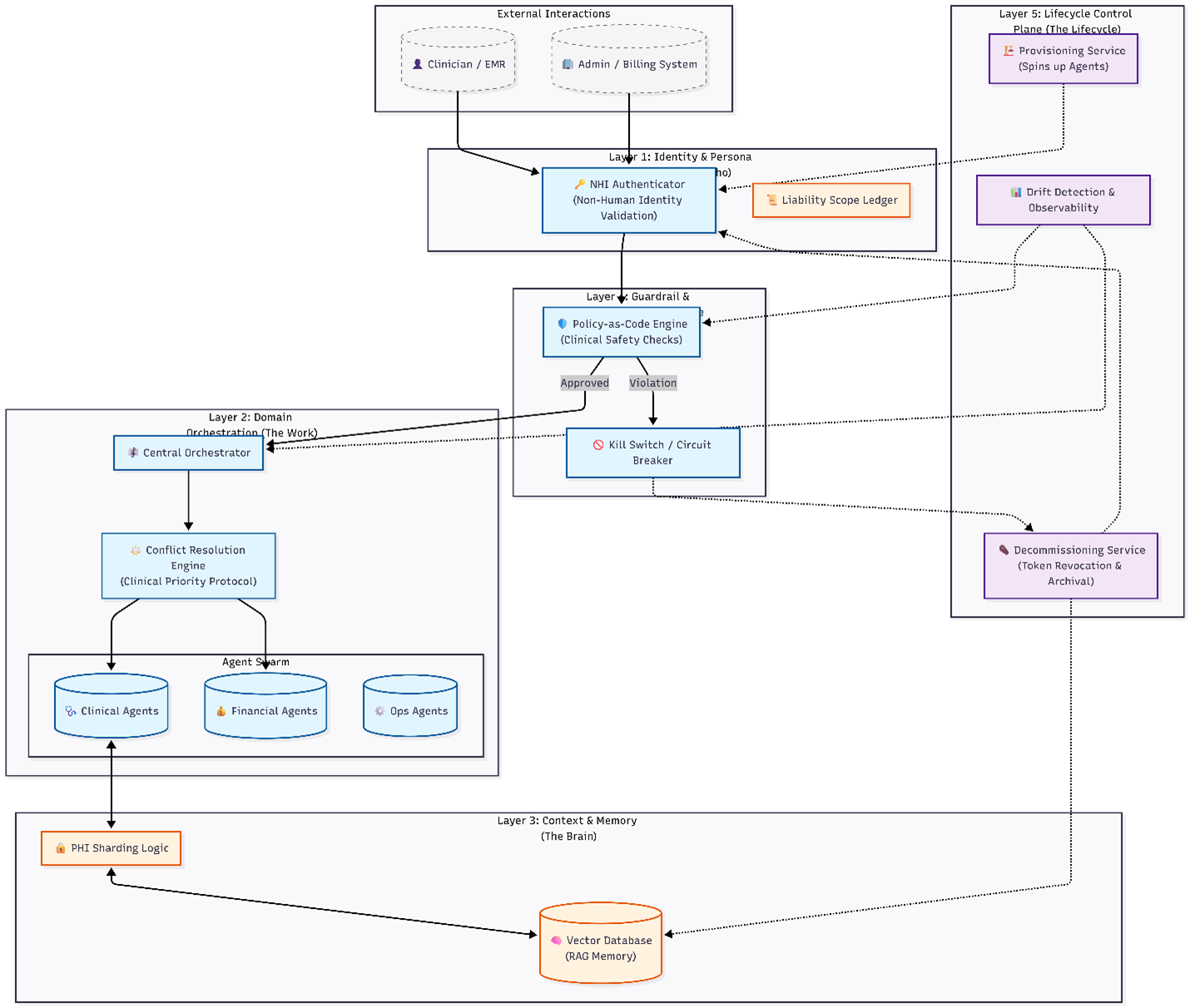}
    \caption{The Unified Agent Lifecycle Management (UALM) Reference Architecture.}
    \label{fig:Picture3}
\end{figure}

Figure \ref{fig:fig9_drawback_analysis} presents the quantified analysis of UALM’s acknowledged limitations. Centralization overhead scaled non-linearly with fleet size (approximately O(n\^1.3)), meaning that the large organization (growing from 22 to approximately 40 agents) faced roughly 4× the governance overhead of the small organization. During simulated control-plane outages (0.5\% monthly probability), incident rates increased approximately 2.8×, indicating a real architectural risk. The autonomy-control trade-off analysis suggested an optimal governance intensity of approximately 74\%, indicating that UALM should not aim for 100\% governance intensity but should preserve some degree of agent autonomy for operational efficiency.

\section{Discussion}
This paper proposes an Agentic AI governance and lifecycle management framework to address gaps in existing AI governance standards and frameworks for healthcare, particularly where autonomy is distributed across multiple interacting agents \cite{Commission2025, Tabassi2023, NIST2024, Habbal2024, Chaffer2025}. The objective is straightforward: reduce agent sprawl by making ownership, permissions, monitoring, and retirement decisions visible and enforceable at the enterprise level.  

Before discussing the framework further, we must recognize its important limitations. First, the simulation uses synthetic scenarios rather than real-world deployment data. While the agent types, event probabilities, and organizational profiles were designed to be realistic, they have not been calibrated against observed healthcare agent fleet operations. Second, the governance ramp-up curves are assumed rather than empirically derived, though the sensitivity analysis demonstrates that results are robust to reasonable variations. Third, the primary limitation of the UALM framework is centralized control, which adds computational overhead and introduces latency. Additionally, defining ground truth in clinical disagreements remains a challenge, requiring human intervention to resolve them. Centralized orchestration and control plane could introduce single points of failure, bottlenecks, and resiliency challenges. Fourth, the single-point-of-failure modeling uses a simplified outage probability rather than a detailed failure mode analysis. Fifth, the simulation does not capture inter-organizational dynamics, vendor relationships, or variability in regulatory enforcement. Sixth, the lead-lag correlations for KPI early warning showed mixed results across organization sizes, suggesting that the proposed KPIs require fleet-size-specific calibration for predictive use. 

Despite the limitations, our proposed framework offers comprehensive coverage of technical depth and functional description to address AI Agent sprawl. Existing standards on healthcare AI are limited to risk management across the lifecycle, covering transparency, human oversight, logging, and post-deployment monitoring \cite{Aboy2024, Kolfschooten2024}, yet remain comparatively underdeveloped on how to engineer these requirements when autonomy is distributed across interacting agents. Analysis and reviews of existing frameworks, such as the EU’s AI Act, available AI governance, and AI TRiSM, cover risk principles but under-specify multi-agent enforcement \cite{Aboy2024, Kolfschooten2024, Cuocolo2025, Batool2025, Finch2025}. A review of AI governance in Table 1 highlights fragmentation between regulatory and operational governance, noting that even complementary risk frameworks can be challenging to translate into actionable, resource-constrained controls \cite{Batool2025, Finch2025}. The AI TRiSM \cite{Habbal2024} frames governance as a system-level program spanning governance, security/privacy, explainability, and operations, while documenting practical barriers such as adversarial threats, compliance burden, and skills gaps \cite{Habbal2024}. Finally, security surveys of LLM-based agents reveal that tool use, memory, and orchestration expand the attack surface, making continuous monitoring and runtime controls central, rather than optional \cite{Tang2026}.

UALM addresses these gaps by translating governance intent into a control plane that is enforceable at runtime and audit-ready by design. Non-human identity provides a clear boundary for agent accountability and revocation. Policy-as-code enables consistent, testable enforcement of permitted actions and tool use \cite{Batool2025, Finch2025}. Policy precedence rules provide a structured way to resolve cross-domain conflicts, with patient safety and privacy constraints taking priority when goals collide, complementing TRiSM’s program-level framing with a concrete coordination safeguard for multi-agent settings \cite{Habbal2024}. PHI segmentation and retention-bound context reduce unnecessary exposure, and drift monitoring, paired with decommissioning, support ongoing lifecycle control, as emphasized in the EU AI Act for regulated healthcare AI \cite{Aboy2024, Cuocolo2025, Tang2026}. This study also provides the empirical evaluation of a comprehensive agentic AI governance framework in a healthcare context. The Monte Carlo simulation demonstrates that UALM produces statistically significant improvements across all seven KPIs, all three organization sizes, and against all comparators (p < 0.001 for all 105 comparisons). The most substantive contribution is the comparison against NIST RMF-Lite, which shows that agent-specific lifecycle governance provides value beyond what generic AI risk management can achieve—particularly in credential revocation automation (d = 6.06), policy-enforced logging (d = 2.95), and PHI minimization (d = 2.83).

The simulation results contextualize UALM’s position relative to the frameworks reviewed in Section 3. Existing standards such as the EU AI Act \cite{eu_ai_act_2024} and NIST AI RMF \cite{NIST2025} cover risk principles but under-specify multi-agent enforcement \cite{laux2021taming}. Our results quantify this gap: NIST RMF-Lite’s generic controls achieved detection rates of 18–55\% across event types, compared to UALM’s 62–74\%. The AI TRiSM framework \cite{gartner_ai_trism} provides program-level framing but lacks agent-specific mechanisms; our Registry-Only condition approximates what inventory-focused approaches can achieve and demonstrates their structural limitations (0\% reaching Level 2 maturity). Agent registration standards like NANDA Index \cite{Raskar2025}, AGNTCY ADS \cite{Muscariello2025}, and Google Agent2Agent \cite{linux_foundation_a2a_2026} address discovery and interoperability but do not provide lifecycle enforcement. UALM’s Layer 1 builds on these foundations while adding the lifecycle layers (2–5) that our simulation shows are essential for detecting drift, conflicts, PHI exposure, and injection events.

The simulation compares four discrete governance conditions and cannot capture the full spectrum of partial implementations that exist in practice. Real healthcare organizations may have heterogeneous governance across departments, which the simulation’s organization-level modeling does not address. Future work should include pilot validation in real healthcare settings to calibrate event probabilities and governance ramp-up trajectories against observed data.

\subsection{Implications}
This study has several managerial implications for healthcare leaders, as it translates Agentic AI governance into an implementation-oriented checklist for enterprise adoption. The value of UALM is managerial as much as technical: it reframes Agentic AI from a set of disconnected procurements into an operating model that can be standardized across the enterprise. The refined maturity model with measurable thresholds enables reliable self-assessment. The strict all-KPI advancement rule creates a high bar: only 18–21\% of UALM runs reached Level 2 within six months. This suggests that even full-framework adoption may require staged implementation and targeted investment. This is by design—the model should identify genuine governance maturity, not provide easy certification. Organizations can use the KPI thresholds to identify specific governance gaps (e.g., credential revocation time is often the binding constraint for Level 2 advancement) and prioritize investments accordingly.

The findings translate into actionable guidance for healthcare CIOs, CISOs, and clinical leaders. First, an agent registry alone is insufficient—organizations must invest in the full governance stack to achieve meaningful risk reduction. Second, UALM requires sustained organizational commitment; the ramp-up sensitivity analysis shows that stalled adoption erases approximately 80\% of benefits. Third, credential revocation and tool-call logging are the highest-leverage investments under the modeled assumptions, as they produce the largest effect sizes across all comparisons. Fourth, organizations should plan for non-linear overhead scaling as agent fleets grow as fleets approach several dozen agents, potentially requiring distributed governance architectures for large academic medical centers.

\section{Conclusions}

This study proposes UALM as a healthcare-specific agent lifecycle governance control plane and uses Monte Carlo simulation to examine its plausibility, sensitivity, and expected operational behavior under alternative governance assumptions. The key contributions are threefold. First, the five-layer UALM architecture addresses governance gaps identified in existing frameworks by providing agent-specific lifecycle controls across identity, orchestration, context, guardrails, and decommissioning. The simulation demonstrates that all five layers contribute meaningfully—each maps to failure types that partial-governance approaches cannot detect. Second, the refined maturity model, with measurable KPI-linked thresholds, enables reliable self-assessment and effectively distinguishes between governance approaches. The strict all-KPI advancement rule ensures that maturity-level assignments reflect genuine governance capability. Third, provided a simulation-based evaluation demonstrates that UALM provides statistically significant improvements over both no-governance baselines and partial-governance alternatives, with the most compelling evidence against NIST RMF-Lite (Cohen’s d = 0.99–6.06 across KPIs). UALM’s advantage is robust across parameter variations but depends on sustained organizational commitment. In conclusion, the risk of agent sprawl in healthcare is not a technology-scaling issue; it is an ethical governance problem, as unmanaged autonomy can undermine safety, accountability, privacy, and trust. By operationalizing Unified Agent Lifecycle Management (UALM), healthcare organizations can move from fragmented, domain-specific agent deployments to a disciplined, managed agent fleet with clear accountability, controlled autonomy, and lifecycle discipline. The objective is not to slow innovation, but to make autonomous agent adoption auditable, accountable, and safe at enterprise scale.

\bibliographystyle{unsrt}  

\bibliography{agentsprawl.bib}

@article{Prakash2026,
   author = {Chandra Prakash and Mary Lind and Elyson De La Cruz},
   doi = {10.62411/jcta.15254},
   issn = {3024-9104},
   issue = {3},
   journal = {Journal of Computing Theories and Applications},
   month = {1},
   pages = {286-301},
   title = {Hybrid Real-time Framework for Detecting Adaptive Prompt Injection Attacks in Large Language Models},
   volume = {3},
   year = {2026}
}

@article{Zohaib2025,
   title={Agentic {AI} in {6G} Software Businesses: A Layered Maturity Model},
   author={Zohaib, Muhammad and Akbar, Muhammad Azeem and Hyrynsalmi, Sami and Khan, Arif Ali},
   journal={arXiv preprint arXiv:2508.03393},
   year={2025},
   url={https://arxiv.org/abs/2508.03393}
}

@inbook{Mulder2023,
   title={A Maturity Model for Collaborative Agents in Human-AI Ecosystems},
   author={Mulder, Wico and Meyer-Vitali, Andr{\'e}},
   booktitle={Collaborative Networks in Digitalization and Society 5.0},
   series={IFIP Advances in Information and Communication Technology},
   volume={688},
   pages={257--269},
   year={2023},
   publisher={Springer},
   doi={10.1007/978-3-031-42622-3_23},
   url={https://link.springer.com/chapter/10.1007/978-3-031-42622-3_23}
}

@article{Schick2023,
   author = {Timo Schick and Jane Dwivedi-Yu and Roberto Dessì and Roberta Raileanu and Maria Lomeli and Luke Zettlemoyer and Nicola Cancedda and Thomas Scialom},
   month = {2},
   title = {Toolformer: Language Models Can Teach Themselves to Use Tools},
   journal={arXiv preprint arXiv:2302.04761},
   year = {2023}
}

@article{Tang2026,
   author = {Yaxin Tang and Yijia Liu and Jiahe Lan and Zheng Yan and Erol Gelenbe},
   doi = {10.1016/j.inffus.2025.103941},
   issn = {15662535},
   journal = {Information Fusion},
   month = {3},
   pages = {103941},
   title = {Security of LLM-based agents regarding attacks, defenses, and applications: A comprehensive survey},
   volume = {127},
   year = {2026}
}

@article{Finch2025,
   author = {William Walter Finch and Marya Butt},
   doi = {10.3390/jcp5040101},
   issn = {2624-800X},
   issue = {4},
   journal = {Journal of Cybersecurity and Privacy},
   month = {11},
   pages = {101},
   title = {Gaps in AI-Compliant Complementary Governance Frameworks’ Suitability (for Low-Capacity Actors), and Structural Asymmetries (in the Compliance Ecosystem)—A Systematic Review},
   volume = {5},
   year = {2025}
}

@article{Batool2025,
   author = {Amna Batool and Didar Zowghi and Muneera Bano},
   doi = {10.1007/s43681-024-00653-w},
   issn = {2730-5953},
   issue = {3},
   journal = {AI and Ethics},
   month = {6},
   pages = {3265-3279},
   title = {AI governance: a systematic literature review},
   volume = {5},
   year = {2025}
}

@article{Cuocolo2025,
   author = {Renato Cuocolo and Diana Bernardini and Daniel Pinto dos Santos and Michail E. Klontzas and Tugba Akinci D’Antonoli and Luís Curvo Semedo and Robin Decoster and Merel Huisman and Elmar Kotter and Luis Martí-Bonmatí and Costin Minoiu and Emanuele Neri and Konstantin Nikolaou and Maija Radzina and Evis Sala and Susan C. Shelmerdine and Laurens Topff and Michelle C. Williams},
   doi = {10.1186/s13244-025-02146-8},
   issn = {1869-4101},
   issue = {1},
   journal = {Insights into Imaging},
   month = {12},
   pages = {275},
   title = {AI medical device post-market surveillance regulations: consensus recommendations by the European Society of Radiology},
   volume = {16},
   year = {2025}
}

@article{Kolfschooten2024,
   author = {Hannah Kolfschooten and Janneke Oirschot},
   doi = {10.1016/j.healthpol.2024.105152},
   issn = {01688510},
   journal = {Health Policy},
   month = {11},
   pages = {105152},
   title = {The EU Artificial Intelligence Act (2024): Implications for healthcare},
   volume = {149},
   year = {2024}
}

@article{Aboy2024,
   author = {Mateo Aboy and Timo Minssen and Effy Vayena},
   doi = {10.1038/s41746-024-01232-3},
   issn = {2398-6352},
   issue = {1},
   journal = {npj Digital Medicine},
   month = {9},
   pages = {237},
   title = {Navigating the EU AI Act: implications for regulated digital medical products},
   volume = {7},
   year = {2024}
}

@misc{NIST2025,
   author = {NIST},
   publisher = {NIST},
   title = {AI RMF Core},
   url = {https://airc.nist.gov/airmf-resources/airmf/5-sec-core/},
   year = {2025}
}

@article{Muscariello2025,
   author = {Luca Muscariello and Vijoy Pandey and Ramiz Polic},
   month = {9},
   title = {The AGNTCY Agent Directory Service: Architecture and Implementation},
   journal={arXiv preprint arXiv:2509.18787},
   year = {2025}
}

@misc{Surapaneni2025,
   author = {Rao Surapaneni and Miku Jha and Michael Vakoc and Todd Segal},
   month = {12},
   title = {Announcing the Agent2Agent Protocol (A2A)},
   url = {https://developers.googleblog.com/en/a2a-a-new-era-of-agent-interoperability/},
   year = {2025}
}

@article{Kasirzadeh2025,
   author = {Atoosa Kasirzadeh and Iason Gabriel},
   month = {4},
   title = {Characterizing AI Agents for Alignment and Governance},
   journal={arXiv preprint arXiv:2504.21848},
   year = {2025}
}

@article{Habbal2024,
   author = {Adib Habbal and Mohamed Khalif Ali and Mustafa Ali Abuzaraida},
   doi = {10.1016/j.eswa.2023.122442},
   issn = {09574174},
   journal = {Expert Systems with Applications},
   month = {4},
   pages = {122442},
   title = {Artificial Intelligence Trust, Risk and Security Management (AI TRiSM): Frameworks, applications, challenges and future research directions},
   volume = {240},
   year = {2024}
}

@article{Chaffer2025,
   author = {Tomer Jordi Chaffer and Charles von Goins and Bayo Okusanya and Dontrail Cotlage and Justin Goldston},
   month = {1},
   title = {Decentralized Governance of Autonomous AI Agents},
   journal={arXiv preprint arXiv:2412.17114},
   year = {2025}
}

@techReport{Nandakumar2025,
   author = {Suhas Nandakumar and Cullen Fluffy Jennings},
   issue = {draft-nandakumar-agent-sd-jwt-01},
   institution = {Internet Engineering Task Force},
   month = {10},
   note = {Work in Progress},
   publisher = {Internet Engineering Task Force},
   title = {Selective Disclosure for Agent Discovery and Identity Management (SD- Agent)},
   url = {https://datatracker.ietf.org/doc/draft-nandakumar-agent-sd-jwt/01/},
   year = {2025}
}

@article{Raskar2025,
   author = {Ramesh Raskar and Pradyumna Chari and John Zinky and Mahesh Lambe and Jared James Grogan and Sichao Wang and Rajesh Ranjan and Rekha Singhal and Shailja Gupta and Robert Lincourt and Raghu Bala and Aditi Joshi and Abhishek Singh and Ayush Chopra and Dimitris Stripelis and Bhuwan B and Sumit Kumar and Maria Gorskikh},
   month = {7},
   title = {Beyond DNS: Unlocking the Internet of AI Agents via the NANDA Index and Verified AgentFacts},
   journal={arXiv preprint arXiv:2507.14263},
   year = {2025}
}

@article{Syros2025,
   author = {Georgios Syros and Anshuman Suri and Jacob Ginesin and Cristina Nita-Rotaru and Alina Oprea},
   month = {8},
   title = {SAGA: A Security Architecture for Governing AI Agentic Systems},
   journal={arXiv preprint arXiv:2504.21034},
   year = {2025}
}

@misc{Commission2025,
   author = {European Commission},
   month = {12},
   title = {AI Act},
   url = {https://digital-strategy.ec.europa.eu/en/policies/regulatory-framework-ai},
   year = {2025}
}

@techReport{NIST2024,
   author = {NIST},
   doi = {10.6028/NIST.AI.600-1},
   institution = {National Institute of Standards and Technology},
   month = {7},
   title = {Artificial intelligence risk management framework :},
   year = {2024}
}

@techReport{Tabassi2023,
   author = {Elham Tabassi},
   doi = {10.6028/NIST.AI.100-1},
   institution = {National Institute of Standards and Technology},
   month = {1},
   title = {Artificial Intelligence Risk Management Framework (AI RMF 1.0)},
   year = {2023}
}

@misc{HHS2009,
   author = {HHS},
   month = {12},
   title = {Summary of the HIPAA Security Rule},
   url = {https://www.hhs.gov/hipaa/for-professionals/security/laws-regulations/index.html?utm_source=chatgpt.com},
   year = {2009}
}

@misc{Nolan2025,
   author = {Beatrice Nolan},
   month = {12},
   title = {An AI-powered coding tool wiped out a software company’s database, then apologized for a ‘catastrophic failure on my part’},
   url = {https://fortune.com/2025/07/23/ai-coding-tool-replit-wiped-database-called-it-a-catastrophic-failure/?utm_source=chatgpt.com},
   year = {2025}
}

@misc{OWASP,
   author = {OWASP},
   title = {OWASP Top 10 for Large Language Model Applications | OWASP Foundation},
   url = {https://owasp.org/www-project-top-10-for-large-language-model-applications/}
}

@misc{OpenAI2025,
   author = {OpenAI},
   month = {12},
   title = {Detecting and reducing scheming in AI models},
   url = {https://openai.com/index/detecting-and-reducing-scheming-in-ai-models/},
   year = {2025}
}

@article{Greenblatt2024,
   author = {Ryan Greenblatt and Carson Denison and Benjamin Wright and Fabien Roger and Monte MacDiarmid and Sam Marks and Johannes Treutlein and Tim Belonax and Jack Chen and David Duvenaud and Akbir Khan and Julian Michael and Sören Mindermann and Ethan Perez and Linda Petrini and Jonathan Uesato and Jared Kaplan and Buck Shlegeris and Samuel R. Bowman and Evan Hubinger},
   month = {12},
   title = {Alignment faking in large language models},
   journal ={arXiv preprint arXiv:2412.14093},
   year = {2024}
}

@article{Liu2025,
   author = {Fei Liu and Yue Niu and Qihua Zhang and Kai Wang and Zheyi Dong and Io Nam Wong and Linling Cheng and Ting Li and Lian Duan and Kun Li and Gen Li and Tai Wa Hou and Manson Fok and Huiyan Luo and Xiangmei Chen and Kang Zhang and Yun Yin},
   doi = {10.1016/j.xcrm.2025.102374},
   issn = {26663791},
   issue = {10},
   journal = {Cell Reports Medicine},
   month = {10},
   pages = {102374},
   title = {A foundational architecture for AI agents in healthcare},
   volume = {6},
   year = {2025}
}

@article{Liu2024,
   author = {Tsai-Ling Liu and Timothy C. Hetherington and Ajay Dharod and Tracey Carroll and Richa Bundy and Hieu Nguyen and Henry E. Bundy and McKenzie Isreal and Andrew McWilliams and Jeffrey A. Cleveland},
   doi = {10.1056/AIoa2400659},
   issn = {2836-9386},
   issue = {12},
   journal = {NEJM AI},
   month = {11},
   title = {Does AI-Powered Clinical Documentation Enhance Clinician Efficiency? A Longitudinal Study},
   volume = {1},
   year = {2024}
}

@misc{Gorenshtein2025,
   author = {Alon Gorenshtein and Mahmud Omar and Benjamin S Glicksberg and Girish N Nadkarni and Eyal Klang},
   doi = {10.1101/2025.08.22.25334232},
   month = {8},
   title = {AI Agents in Clinical Medicine: A Systematic Review},
   year = {2025}
}

@unpublished{Banerjie2025,
   author = {Shruti Banerjie and Yuxin Zhu and Isaac Freeman and Julyssa Villa Machado and Abdulaziz Ahmed and Abeed Sarker and Mohammed Al-Garadi},
   doi = {10.36227/techrxiv.176238073.31262603/v1},
   institution = {Techrxiv},
   month = {11},
   title = {Agentic AI in Healthcare: A Comprehensive Survey of Foundations, Taxonomy, and Applications},
   year = {2025},
   note = {Unpublished manuscript}
}

@article{Zhang2023,
   author = {Peng Zhang and Maged N. Kamel Boulos},
   doi = {10.3390/fi15090286},
   issn = {1999-5903},
   issue = {9},
   journal = {Future Internet},
   month = {8},
   pages = {286},
   title = {Generative AI in Medicine and Healthcare: Promises, Opportunities and Challenges},
   volume = {15},
   year = {2023}
}

@article{Shokrollahi2025,
   author = {Yasin Shokrollahi and Jose Colmenarez and Wenxi Liu and Sahar Yarmohammadtoosky and Matthew M. Nikahd and Pengfei Dong and Xianqi Li and Linxia Gu},
   month = {8},
   title = {Recent Advances in Generative AI for Healthcare Applications},
   journal   = {Journal of Imaging Informatics in Medicine},
   volume    = {39},
   year      = {2026},
   doi       = {10.1007/s10278-026-01908-0}
}

@article{Karunanayake2025,
   author = {Nalan Karunanayake},
   doi = {10.1016/j.infoh.2025.03.001},
   issn = {29499534},
   issue = {2},
   journal = {Informatics and Health},
   month = {9},
   pages = {73-83},
   title = {Next-generation agentic AI for transforming healthcare},
   volume = {2},
   year = {2025}
}

@article{Brodeur2025,
   author = {Peter G. Brodeur and Thomas A. Buckley and Zahir Kanjee and Ethan Goh and Evelyn Bin Ling and Priyank Jain and Stephanie Cabral and Raja-Elie Abdulnour and Adrian D. Haimovich and Jason A. Freed and Andrew Olson and Daniel J. Morgan and Jason Hom and Robert Gallo and Liam G. McCoy and Haadi Mombini and Christopher Lucas and Misha Fotoohi and Matthew Gwiazdon and Daniele Restifo and Daniel Restrepo and Eric Horvitz and Jonathan Chen and Arjun K. Manrai and Adam Rodman},
   month = {6},
   journal   = {Science},
   title = {Superhuman performance of a large language model on the reasoning tasks of a physician},
   volume    = {392},
   number    = {6797},
   pages     = {524--527},
   year      = {2025},
   doi       = {10.1126/science.adz4433}
}

@book{rubinstein2017simulation,
  title={Simulation and the Monte Carlo Method},
  author={Rubinstein, Reuven Y. and Kroese, Dirk P.},
  edition={3rd},
  year={2017},
  publisher={John Wiley \& Sons},
  address={Hoboken, NJ},
  isbn={978-1118632161}
}

@misc{eu_ai_act_2024,
  author       = {{European Parliament} and {Council of the European Union}},
  title        = {Regulation ({EU}) 2024/1689 of the {European Parliament} and of the {Council} of 13 {June} 2024 laying down harmonised rules on artificial intelligence ({Artificial Intelligence Act}) and amending certain {Union} legislative acts},
  journal      = {Official Journal of the European Union},
  volume       = {67},
  number       = {L},
  year         = {2024},
  month        = jul,
  day          = {12},
  url          = {https://eur-lex.europa.eu/eli/reg/2024/1689/oj},
  note         = {OJ L, 2024/1689, 12.7.2024}
}

@misc{gartner_ai_trism,
  author       = {{Gartner}},
  title        = {Tackling Trust, Risk and Security in {AI} Models ({AI TRiSM})},
  howpublished = {Gartner Research},
  year         = {2024},
  url          = {https://www.gartner.com/en/articles/ai-trust-and-ai-risk},
  note         = {Accessed: May 2026}
}

@article{laux2021taming,
  title        = {Taming the few: {Platform} regulation, independent audits, and the risks of capture created by the {DMA} and {DSA}},
  author       = {Laux, Johann and Wachter, Sandra and Mittelstadt, Brent},
  journal      = {Computer Law \& Security Review},
  volume       = {43},
  pages        = {105613},
  year         = {2021},
  publisher    = {Elsevier},
  doi          = {10.1016/j.clsr.2021.105613},
  issn         = {0267-3649},
  url          = {https://doi.org/10.1016/j.clsr.2021.105613}
}

@techreport{linux_foundation_a2a_2026,
  author       = {{Linux Foundation} and {Google}},
  title        = {Agent-to-Agent ({A2A}) Protocol Specification},
  institution  = {Linux Foundation},
  type         = {Technical Specification},
  version      = {v0.3.0 RC1},
  year         = {2026},
  month        = mar,
  url          = {https://a2a-protocol.org/},
  note         = {Available at the official repository: https://github.com/a2aproject/A2A}
}

\end{document}